%% file: main.tex
\documentclass[11pt,a4paper,copyright]{google}

\usepackage[authoryear,sort&compress,round]{natbib}
\input{sec/preamble}

\input{sec/abstract}

\begin{document}
\maketitle
\input{sec/overview}
\Needspace{6\baselineskip}
\input{sec/introduction}
\input{sec/related_work}
\input{sec/method}
\input{sec/experiments}
\Needspace{8\baselineskip}
\input{sec/conclusion}
\FloatBarrier

\begingroup
\hypersetup{urlcolor=black}
\renewcommand{\bibfont}{\small}
\setlength{\bibsep}{3pt plus 1pt minus 1pt}
\bibliography{references}
\endgroup
\input{sec/appendix}
\end{document}

%% file: sec/preamble.tex
\usepackage{spac_frontmatter}
\usepackage{bm}
\usepackage{algorithm,algorithmic}
\usepackage{adjustbox}
\usepackage{placeins}
\usepackage{wrapfig}
\usepackage{xurl}

\newcommand{\pubvenue}[1]{{\normalfont\scriptsize\textcolor{black!55}{(#1)}}}

\def\eqref#1{(\ref{#1})}

\def\plaineqref#1{\ref{#1}}

\def\1{\bm{1}}

\newcommand{\firstpara}[1]{\noindent\textbf{#1}}
\newcommand{\para}[1]{\smallskip\noindent\textbf{#1}}

\newcommand{\papertitle}{From Dense to Sparse: Morton-Ordered Sparse Mapping for Efficient 3D Mesh Generation}
\title{\papertitle}

\renewcommand{\Affilfont}{\normalfont\fontsize{10.5}{13}\selectfont}
\author[1]{Junjie~Yang}
\author[2]{Rui~Qian}
\author[3]{Yijiang~Li}
\author[4]{Jinyi~Xu}
\author[5]{Yuhao~Yan}
\author[6]{Xinjian~Yi}
\author[7]{Yanfeng~Li}
\author[2]{Mingxuan~Li}
\author[8]{Xiang~Wan}
\author[9]{Jicong~Fan}
\author[10]{Siqi~Liu}
\author[11]{Ahmed~Elazab}
\author[8\,\CorrespondingAuthorSymbol]{Changmiao~Wang}

\affil[1]{\mbox{South China University of Technology}}
\affil[2]{\mbox{Fudan University}}
\affil[3]{\mbox{University of California, San Diego}}
\affil[4]{\mbox{Beihang University}}
\affil[5]{\mbox{Zhejiang University}}
\affil[6]{\mbox{Southeast University}}
\affil[7]{\mbox{Macao Polytechnic University}}
\affil[8]{\mbox{Shenzhen Research Institute of Big Data}}
\affil[9]{\mbox{The Chinese University of Hong Kong Shenzhen}}
\affil[10]{\mbox{National Health Institute of Big Data, Shenzhen}}
\affil[11]{\mbox{Tsinghua University}}

\definecolor{spacalgblue}{RGB}{0,102,190}
\definecolor{spacalgred}{RGB}{205,0,0}
\definecolor{spacalgorange}{RGB}{225,115,0}
\newcommand{\spacalgcomment}[2]{%
  \unskip\nobreak\hspace{0.5em}\hfill%
  \makebox[58pt][l]{\fontsize{8}{9.5}\selectfont%
    \textcolor{#1}{$\triangleright$}\hfill#2}%
  \ignorespaces
}

\definecolor{royalblue}{RGB}{65,105,225}
\usepackage[colorlinks=true,
            linkcolor=red,
            urlcolor=royalblue,
            citecolor=brown,
            anchorcolor=blue]{hyperref}

\renewcommand{\absfont}{\normalfont\linespread{1.2}\fontsize{11}{12}\selectfont}

%% file: sec/abstract.tex
\begin{abstract}

Unified 3D understanding and generation require compact sequences preserving spatial structure. Dense grids yield long sequences, while serialization can separate spatial neighbors. We propose \textbf{S}parse \textbf{P}roximity \textbf{A}ware \textbf{C}ompression (SPAC), a shared sparse SDF vocabulary combining sparse encoding with reversible spatial merging. A Sparse SDF VQ-VAE quantizes near-surface features while retaining explicit coordinates, using reconstruction supervision for surface fidelity and detail. 3D Byte-Pair Encoding (3D-BPE) merges recurring face-adjacent patterns using relative offsets before serialization. Macro tokens expand exactly into leaf tokens and coordinates. After Morton serialization, a shared LLM reads geometry for captioning and predicts tokens that decode into meshes for text-conditioned generation. With fixed geometry, architecture, and band width in voxels, sparse encoder cost grows approximately quadratically with resolution. 3D-BPE reduces mean complete LLM sequence length by 45.4 percent, including coordinates, text, and control tokens. SPAC achieves strong results in both 3D captioning and text-conditioned mesh generation. Ablations show gains over leaf tokens and Morton 1D-BPE in both tasks. 

\end{abstract}

%% file: sec/overview.tex
\begin{figure}[!htbp]
\centering
\includegraphics[width=\linewidth]{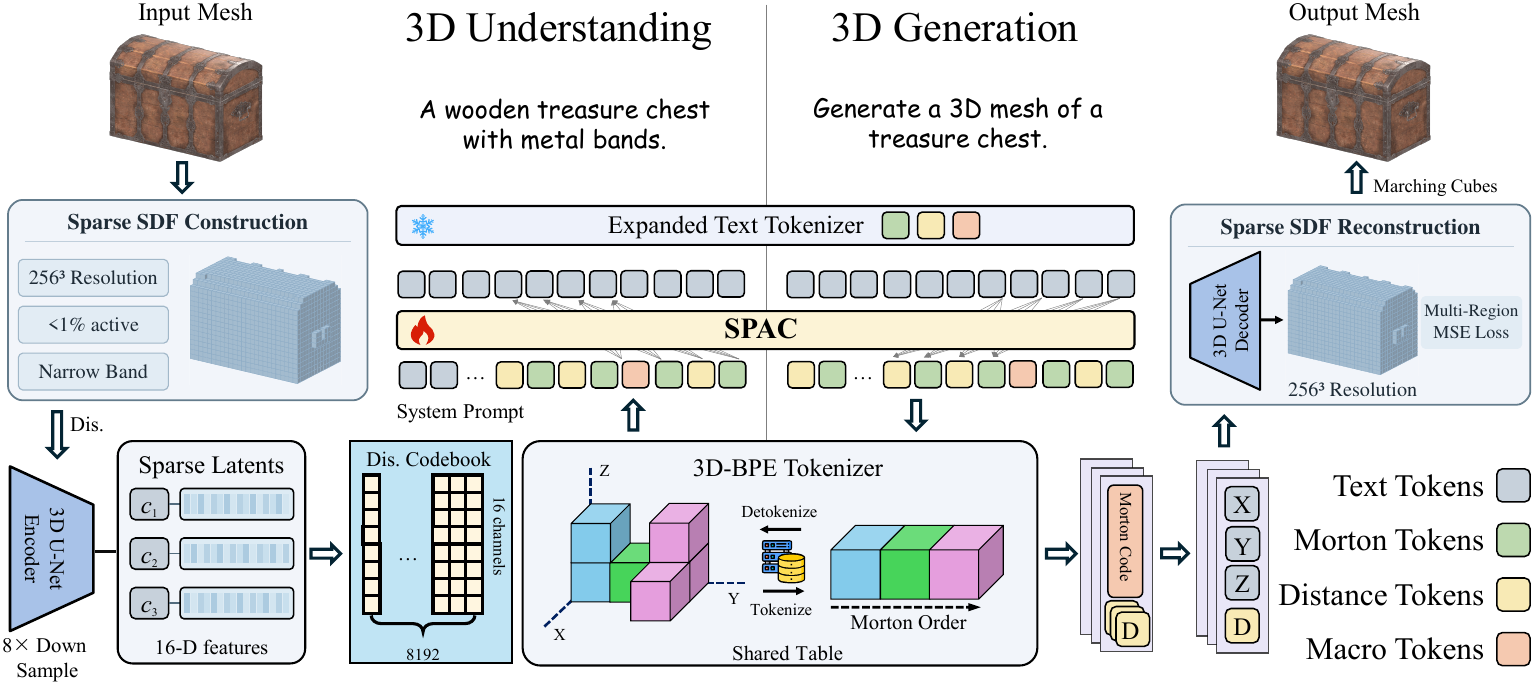}
\caption{\textbf{Overview of SPAC.} \textit{(1) Sparse Encoding:} A trained Sparse SDF VQ-VAE converts near-surface geometry into discrete leaf tokens with explicit coordinates. \textit{(2) 3D-BPE:} Reversible, offset-conditioned merges combine recurring face-adjacent patterns into macro tokens. \textit{(3) Unified Modeling:} The LLM tokenizer is extended with Morton, distance, and macro tokens. After Morton serialization, a shared autoregressive LLM reads geometry tokens for understanding and predicts them for generation. \textit{(4) Reconstruction:} Predicted macro tokens expand into leaf tokens, which the VQ-VAE decodes into a sparse SDF for mesh extraction with Marching Cubes.}
\label{fig:sparc}
\end{figure}

%% file: sec/introduction.tex
\section{Introduction} \label{sec:intro}

Autoregressive next-token prediction offers large language models (LLMs) a common interface for understanding and generation~\citep{brown2020language}. Models such as Emu3~\citep{wang2024emu3}, Chameleon~\citep{aghajanyan2024chameleon}, and Transfusion~\citep{zhou2024transfusion} extend this interface to visual content through discrete or hybrid representations. For 3D, this requires a geometry vocabulary that keeps sequences compact while retaining spatial structure for decoding. Such a vocabulary must serve both directions: the model should interpret geometric structure when describing an object and produce decodable geometry when conditioned on text. This makes token design central to unification, because the same sequence determines what the language model observes and what the geometric decoder can reconstruct. A useful representation must therefore balance sequence length, spatial organization, and recoverability.

LLaMA-Mesh~\citep{wang2024llamamesh} and ShapeLLM-Omni~\citep{ye2025shapellmomni} unify generation and understanding through textual meshes and quantized voxels. XCube~\citep{ren2024xcube}, Direct3D-S2~\citep{wu2025direct3ds2}, and TRELLIS~\citep{xiang2025structured} concentrate computation near surfaces, typically using diffusion or rectified flow for generation. A compact shared vocabulary must address three issues. First, dense voxel token counts grow cubically with resolution, so increasing geometric detail also increases the sequence the LLM must process. Restricting encoding to a surface band addresses spatial occupancy, but does not remove repeated local patterns among active sites. Second, serialization can separate spatial neighbors: tokens that form one geometric structure need not be adjacent in a one-dimensional sequence. Compression based only on sequence proximity may therefore overlook recurring 3D patterns. Third, implicit coordinates tie geometry to sequence positions, complicating variable-size token composition. Merging such tokens must also preserve the positions needed to reconstruct their constituent geometry. These issues are coupled. A short sequence is useful only if its tokens still identify where geometry lies and how local parts fit together. Combining sparsity with compression therefore requires a representation that exposes spatial relations to the tokenizer and retains a defined route back to the original quantized geometry. These constraints motivate sparse encoding, explicit coordinates, and reversible spatial merging.

We propose \textbf{S}parse \textbf{P}roximity \textbf{A}ware \textbf{C}ompression (SPAC), which combines sparse encoding with reversible spatial merging. A Sparse SDF VQ-VAE quantizes near-surface latent features into tokens while retaining explicit coordinates. Its reconstruction objective balances surface fidelity and fine geometric detail. For a fixed surface and band width in voxel units, the number of active sites grows approximately quadratically with resolution. Explicit coordinates also make neighborhood relations available to the compression stage without depending on a chosen serialization. Sparsity reduces the number of encoded locations, but recurring local patterns remain. We introduce 3D Byte-Pair Encoding (3D-BPE) to learn merges on the face adjacency graph using relative offsets. Merging before serialization captures recurring spatial patterns without requiring their tokens to be consecutive in a sequence. Each resulting macro token expands exactly into its constituent leaf tokens and coordinates. Relative offsets record how constituent patterns are placed, allowing a learned merge to be reused across translations. This exact recovery preserves the quantized coordinate--token mapping, while the VQ-VAE determines surface reconstruction through its codebook and decoder.

After Morton serialization, a shared LLM reads geometry tokens for captioning and predicts them for text-conditioned generation. Predicted macro tokens expand into leaf tokens, the VQ-VAE reconstructs a sparse SDF, and Marching Cubes extracts the mesh directly. Both tasks consequently share geometric units, and generation follows a defined path from language-model predictions to a mesh. Experiments support favorable sparse encoder scaling. 3D-BPE reduces mean complete LLM sequence length by 45.4 percent, including coordinates, text, and control tokens. Among evaluated models, SPAC leads all five captioning metrics and achieves strong mesh generation results. Ablations show gains in both tasks over leaf tokens and Morton 1D-BPE. We report encoder costs separately from LLM sequence lengths so that savings at these two stages can be assessed in their respective settings. Our contributions are threefold.
\begin{enumerate}
    \interlinepenalty=10000
    \item We introduce SPAC, a unified autoregressive model that uses shared Sparse SDF VQ-VAE tokens with explicit coordinates for captioning and text-conditioned mesh generation within one autoregressive LLM.
    \item We present 3D-BPE, which learns reversible merges on the face adjacency graph using relative offsets, forming a compact and reusable vocabulary of recurring local spatial patterns before serialization.
    \item We demonstrate favorable sparse encoder scaling, shorter complete LLM sequences, and gains in both captioning and generation over leaf tokens and Morton 1D-BPE.
\end{enumerate}

%% file: sec/related_work.tex
\section{Related Work} \label{sec:related}

\firstpara{Discrete 3D Representations and Autoregressive Generation.}
Classical implicit and hybrid representations provide continuous geometric priors~\citep{mildenhall2021nerf,shen2021dmtet,zhang2023shape2vecset} but do not directly support discrete sequence modeling. VQ tokenizers~\citep{vandenoord2017neural,razavi2019generating,esser2021taming} bridge this gap by quantizing geometry for decoder-only Transformers. MeshGPT~\citep{siddiqui2024meshgpt}, MeshAnything~\citep{yiwen2024meshanything,yiwen2024meshanythingv2}, and MeshXL~\citep{sijin2024meshxl} use face-, vertex-, or coordinate-based sequences, enabling autoregressive synthesis but suffering from long sequences and ordering sensitivity at high resolutions. AutoSDF~\citep{mittal2022autosdf} learns an autoregressive prior over quantized SDF grids, while ShapeFormer~\citep{yan2022shapeformer} uses sparse quantized implicit codes for shape completion. BPT~\citep{weng2025bpt} compresses explicit mesh sequences via block indexing and patch aggregation; FreeMesh~\citep{liu2025freemesh} rearranges and merges frequent coordinate patterns. MeshMosaic~\citep{xu2026meshmosaic} assembles autoregressively generated mesh patches, and MeshWeaver~\citep{xu2026meshweaver} uses sparse voxels to guide vertex prediction. Sparse volumetric methods, including XCube~\citep{ren2024xcube}, Direct3D-S2~\citep{wu2025direct3ds2}, Sparc3D~\citep{li2025sparc3d}, TRELLIS~\citep{xiang2025structured}, and O-Voxel~\citep{xiang2025nativecompactstructuredlatents}, concentrate computation near occupied or surface regions, enabling higher resolutions. Their generative stages typically use diffusion or rectified flow. Other routes include diffusion-guided optimization~\citep{poole2023dreamfusion,lin2023magic3d}, diffusion-based view synthesis~\citep{liu2023zero123,long2024wonder3d}, feed-forward reconstruction~\citep{xu2024instantmesh,hong2024lrm}, and latent diffusion~\citep{zhao2023michelangelo,wang2022rodin,zhang2024clay,li2025craftsman3d,yang2024hunyuan3d}. SPAC combines sparse SDF codes with reversible, offset-conditioned merging on the 3D face-adjacency graph to form a compact geometry vocabulary.

\firstpara{Multimodal Large Language Models for 3D.} Building on vision--language alignment~\citep{radford2021learning,alayrac2022flamingo,liu2023llava,dai2023instructblip,chen2024internvl}, 3D-LLM~\citep{hong2023injecting3d}, PointLLM~\citep{xu2023pointllm}, and ShapeLLM~\citep{qi2024shapellm} inject 3D features into LLMs for understanding. SAR3D~\citep{chen2025sar3d} uses multi-scale triplane codes for next-scale generation and a separately fine-tuned LLM for understanding. Recent work moves toward unified understanding and generation: ShapeLLM-Omni~\citep{ye2025shapellmomni} incorporates 3D VQ tokens for generative and discriminative tasks, LLaMA-Mesh~\citep{wang2024llamamesh} represents meshes as UTF-8 text, and EVA01~\citep{yang2026eva01unifiednative3d} uses specialized experts to unify 3D understanding, generation, and editing. ShapeLLM-Omni also describes a rectified-flow route from voxels to high-quality meshes~\citep{ye2025shapellmomni}. SPAC follows the unified modeling direction with sparse SDF tokens that decode directly into mesh geometry.

\firstpara{Long-Sequence Modeling and Compression.}
BPE~\citep{sennrich2016neural} merges frequent adjacent token pairs. SentencePiece~\citep{kudo2018sentencepiece} supports BPE and unigram tokenization, while visual BPE~\citep{zhang2025pixelstotokens} extends merging to quantized image patches. IO-aware exact kernels~\citep{dao2022flashattention}, hardware-aligned sparse attention~\citep{deepseek2025nsa}, and linear-time diffusion backbones~\citep{xie2025sana} reduce sequence-processing costs; token compression instead targets representation length. Uni-3DAR~\citep{lu2025uni3dar} compresses spatial tokens using octrees. 3D-BPE reduces sequence length before LLM modeling while preserving explicit spatial structure for exact expansion into leaf tokens and coordinates.

%% file: sec/method.tex
\section{Method} \label{sec:method}

SPAC uses a shared sparse SDF vocabulary for 3D understanding and generation, combining reversible spatial compression with autoregressive modeling.

\subsection{Preliminary: Dense Autoregressive Modeling via Sequence Flattening}
\label{sec:preliminary_dense_ar}

\begingroup
\predisplaypenalty=0
\firstpara{Formulations.}
Autoregressive modeling represents 3D shapes by quantizing geometry into a token grid and flattening it into a 1D sequence. A 3D VQ-VAE~\citep{vandenoord2017neural} encodes a shape into a discrete token grid $q \in [V]^{R \times R \times R}$, where $R$ is the spatial resolution, and a fixed function $\pi: [R]^3 \to [T]$ flattens $q$ into a 1D sequence $x = (x_1, \dots, x_T)$ with $T = R^3$. The likelihood factorizes as
\begin{align}
    p(x_1, \dots, x_T) = \prod_{t=1}^{T} p(x_t \mid x_1, \dots, x_{t-1}).
    \label{eq:ar_factorization}
\end{align}
\endgroup
This chain-rule factorization models tokens through their sequence prefixes. In a dense representation, each coordinate is recoverable from its index through $\pi^{-1}$, while spatial locality remains implicit in the serialization order.

\para{Limitations of Dense Autoregressive Modeling.}
Dense flattening presents three challenges:
\begin{enumerate}[label=\arabic*),topsep=1pt,itemsep=2pt,leftmargin=20pt]
\item \textbf{Spatial Locality.}
A voxel's lattice face neighbors are
\begin{align}
   \mathcal{N}_{\mathrm{face}}(c)=\{c'\in\mathbb Z^3\colon \|c'-c\|_1=1\}.
   \label{eq:face_neighbors}
\end{align}
Flattening can place these neighbors far apart, requiring the model to learn local geometry through long-range sequence dependencies.

\item \textbf{Implicit Coordinates.}
Coordinates are determined by sequence indices through $\pi^{-1}$, tying token positions and sequence length to the dense grid rather than allowing variable-length sparse representations.

\item \textbf{Computational Cost.}
The dense sequence contains $T=R^3$ tokens, making $O(T^2)$ global attention and step-by-step generation costly at high resolutions.
\end{enumerate}

\begingroup
\interlinepenalty=10000
\para{From Dense Grids to Sparse Geometry.}
Meaningful SDF variation concentrates near the object surface. A sparse coordinate--token mapping allocates tokens to this narrow band, retains explicit positions, and makes 3D adjacency directly computable.
\par
\endgroup

\subsection{Sparse SDF VQ-VAE and Coordinate Mapping}
\label{sec:sparse_mapping}

\firstpara{Reformulation.} A sparse quantized SDF grid provides discrete codes for language modeling and explicit lattice coordinates for face-adjacency queries and reversible merging.
We represent a 3D shape as a sparse coordinate--token mapping:
\begin{align}
    \mathcal{M}= \{(c_i,k_i)\}_{i=1}^{M},
    \qquad
    c_i\in\mathbb{Z}^{3},\quad k_i\in[V],
    \label{eq:sparse_map}
\end{align}
where $c_i$ is an active latent coordinate, $k_i$ is its discrete geometry token, and $M$ is the number of active latent locations. Explicit coordinates define face adjacency for 3D-BPE, while Morton serialization supplies a locality-aware order for the LLM.

\para{Sparse SDF Encoding and Quantization.}
We construct a Sparse SDF VQ-VAE. For a normalized shape, the SDF is computed on a grid of resolution $R{=}256$. We retain a wider band for reconstruction supervision and use its inner near-surface subset as the encoder input:
\begin{align}
    \tau_w &= \frac{\eta_w}{R}, \qquad
    \bar{s}(\mathbf{x})
    = \operatorname{clip}\!\left(\frac{s(\mathbf{x})}{\tau_w},-1,1\right),
    \notag\\*[-2pt]
    \mathcal{V}_w
    &= \bigl\{(\mathbf{x},\bar{s}(\mathbf{x})) \mid
    |s(\mathbf{x})|<\tau_w\bigr\},
    \notag\\*[-2pt]
    \mathcal{V}_e
    &= \bigl\{(\mathbf{x},\bar{s}(\mathbf{x}))\in\mathcal{V}_w
    \mid |\bar{s}(\mathbf{x})|\leq \eta_e/\eta_w\bigr\},
    \label{eq:sparse_sdf}
\end{align}
\begingroup
\predisplaypenalty=0
where $\eta_w{=}4.0$ and $\eta_e{=}0.5$. Thus, the encoder focuses on the most informative surface neighborhood, while the wider band supplies contextual SDF supervision. The sparse 3D U-Net from Direct3D-S2~\citep{wu2025direct3ds2} encodes the SDF into an $8{\times}$ downsampled $32^3$ latent grid with $C_z{=}16$ channels. Each active vector $\mathbf{z}_i\in\mathbb{R}^{C_z}$ is quantized using codebook $\mathbf{E}\in\mathbb{R}^{K\times C_z}$, where $K=V=v_0=8192$:
\begin{align}
k_i^*
&= \arg\min_{k\in[K]}
   \|\mathbf{z}_i-\mathbf{e}_k\|_2,
   \notag\\*[-2pt]
\hat{\mathbf{z}}_i
&= \mathbf{z}_i
 + \mathrm{sg}\bigl(\mathbf{e}_{k_i^*}-\mathbf{z}_i\bigr),
\label{eq:vq}
\end{align}
\endgroup
where $\mathrm{sg}(\cdot)$ is the stop-gradient operator. This yields one discrete leaf token for each active location of the $32^3$ latent grid, together with its explicit coordinates.

\begingroup
\displaywidowpenalty=10000
\para{VQ-VAE Training.}
Training proceeds in two stages. Stage 1 freezes the pretrained encoder--decoder and trains only the codebook using exponential moving average (EMA) updates and K-means reinitialization to prevent collapse. Stage 2 fine-tunes the encoder--decoder while continuing EMA codebook updates. To preserve both surface fidelity and fine geometric structure, reconstruction supervision is stratified into regular surface samples, sharp geometric features, and contextual samples in the wider band. Let $\mathcal{G}=\{\mathrm{surf},\mathrm{feat},\mathrm{ctx}\}$ \mbox{denote} the regions, $\Omega_r$ their supervised locations, and $\tilde{s}$ the reconstructed normalized SDF. The objective is
\begin{align}
    \mathcal{L}_{\mathrm{recon}}
    &= \sum_{r\in\mathcal{G}}
    \frac{\lambda_r}{|\Omega_r|}
    \sum_{\mathbf{x}\in\Omega_r}
    \left(\bar{s}(\mathbf{x})-\tilde{s}(\mathbf{x})\right)^2,
    \\[2pt]
    \mathcal{L}_{\mathrm{commit}}
    &= \frac{1}{M}
    \sum_{i=1}^{M}
    \left\|
    \mathbf{z}_i
    -\mathrm{sg}\!\left(\mathbf{e}_{k_i^*}\right)
    \right\|_2^2,
    \\[2pt]
    \mathcal{L}
    &= \mathcal{L}_{\mathrm{recon}}
    + \lambda_c\,\mathcal{L}_{\mathrm{commit}}.
    \label{eq:vae_loss}
\end{align}
\endgroup
We set $\lambda_{\mathrm{surf}}=\lambda_{\mathrm{feat}}=\lambda_{\mathrm{ctx}}=1$ and $\lambda_c=0.25$. Normalizing each component independently prevents sparse geometric features from being overwhelmed by regular surface samples. During training, the wider band also defines valid sparse support at successive decoder scales, constraining reconstruction to geometrically relevant voxels; predictions outside this support are excluded from reconstruction supervision used for decoder training.

\subsection{3D Byte-Pair Encoding: Compressing While Preserving Structure}
\label{sec:3dbpe}

\begin{figure}[!htb]
\centering
\includegraphics[width=\linewidth]{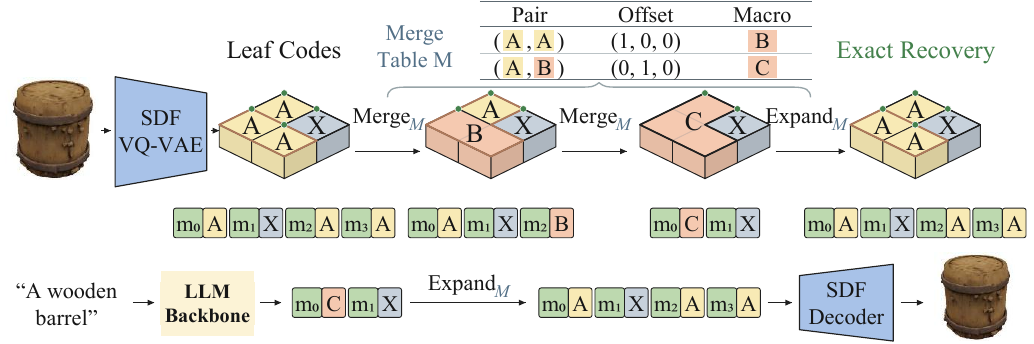}
\caption{\textbf{Spatial compression with 3D-BPE.} Subscripts $M$ denote lookup in the shared merge table, learned offline and then fixed. The local example illustrates face-adjacent merging and exact recovery of leaf codes and coordinates. Below, SPAC predicts coordinate--macro pairs, which are expanded using the same table and decoded into a mesh.}
\label{fig:spac_dataflow}
\end{figure}

\firstpara{Motivation.}
Sparse mapping reduces tokens to surface scale, but recurring local structures still span multiple leaf tokens. Inspired by BPE for text~\citep{sennrich2016neural} and images~\citep{zhang2025pixelstotokens}, 3D-BPE groups these structures into macro tokens, providing compact spatial prediction units for the LLM.

\Needspace{390pt}
\begin{wrapfloat}{algorithm}[25]{r}{0.49\linewidth}
\footnotesize
\raggedright
\setlength{\algorithmicindent}{0.65em}
\algsetup{linenosize=\scriptsize}
\setboolean{ALC@noend}{true}
\renewcommand{\spacalgcomment}[2]{%
  \hspace{0.35em plus 0.2em}\hfill
  \mbox{\scriptsize\textcolor{#1}{$\triangleright$}~#2}}
\captionsetup{position=top,font=footnotesize,labelfont=bf,labelsep=period,
  justification=raggedright,skip=3pt}
\caption{3D-BPE Learning, Encoding and Decoding}
\label{alg:3d-bpe-train}
\begin{algorithmic}[1]
\linespread{0.97}\selectfont
\renewcommand{\algorithmicrequire}{\textbf{Input.}}
\renewcommand{\algorithmicensure}{\textbf{Output.}}
\renewcommand{\algorithmiccomment}[1]{#1}
\REQUIRE Corpus $\mathcal D=\{\mathcal M_i\}$, new shape $\mathcal M$, vocabulary size $v_0$, merge budget $m$, threshold $\tau$
\ENSURE Rules $\mathcal M_{\mathrm{BPE}}$, vocabulary $\mathcal V$, patterns $\mathcal P$, sequence $\mathcal S$, recovered mapping $\widehat{\mathcal M}$
\STATE $\mathcal V\gets\{0,\ldots,v_0-1\}$, $\mathcal M_{\mathrm{BPE}}\gets[\,]$
\STATE $\mathcal P_k\gets\{(\mathbf 0,k)\}$ for $k\in\mathcal V$,\newline
       $\mathcal B_i\gets\textsc{InitBlocks}(\mathcal M_i)$ for $\mathcal M_i\in\mathcal D$
\FOR{$t=1$ \TO $m$}
    \STATE \textcolor{spacalgblue}{$f\gets\textsc{CountFacePairs}(\{\mathcal B_i\})$}\spacalgcomment{spacalgblue}{3D adjacency}
    \STATE \textbf{if} $f$ is empty \textbf{then break}
    \STATE $(a,b,r)\gets\arg\max_p f(p)$, break ties lexicographically
    \STATE \textbf{if} \textcolor{spacalgred}{$f(a,b,r)<\tau$} \textbf{then break}\spacalgcomment{spacalgred}{Freq. cutoff}
    \STATE $Q\gets\mathcal P_a\cup\{(r+d,k)\mid(d,k)\in\mathcal P_b\}$ \label{alg:bpe-pattern-start}
    \STATE $u\gets|\mathcal V|$, $s\gets\min_{\mathrm{lex}}\{d\mid(d,k)\in Q\}$\label{alg:bpe-anchor}\spacalgcomment{spacalgorange}{New anchor}
    \STATE \textcolor{spacalgorange}{$\mathcal P_u\gets\{(d-s,k)\mid(d,k)\in Q\}$}, $\mathcal V\gets\mathcal V\cup\{u\}$ \label{alg:bpe-pattern-end}
    \FOR{each shape $i$}
        \STATE $\mathcal C\gets\textsc{MatchPairs}(\mathcal B_i,a,b,r)$, $\mathcal U\gets\emptyset$ \label{alg:bpe-replace-start}
        \FOR{each $(A,B)$ in $\mathcal C$}
            \STATE \textbf{if} $A\in\mathcal U$ or $B\in\mathcal U$ \textbf{then continue}
            \STATE $N\gets(u,\alpha_A+s,G_A\cup G_B)$, $\mathcal U\gets\mathcal U\cup\{A,B\}$
            \STATE $\mathcal B_i\gets(\mathcal B_i\setminus\{A,B\})\cup\{N\}$\label{alg:bpe-replace-end}\spacalgcomment{spacalgorange}{Greedy merge}
        \ENDFOR
    \ENDFOR
    \STATE Append \textcolor{spacalgblue}{$(u,a,b,r)$} to $\mathcal M_{\mathrm{BPE}}$\spacalgcomment{spacalgblue}{Rule storage}
\ENDFOR
\STATE $\mathcal B\gets\textsc{InitBlocks}(\mathcal M)$, $\widehat{\mathcal M}\gets\emptyset$ \label{alg:bpe-encode-start}
\FOR{each $(u,a,b,r)$ in $\mathcal M_{\mathrm{BPE}}$ in learned order}
    \STATE $\mathcal B\gets\textsc{ApplyRule}(\mathcal B,u,a,b,r,\mathcal P)$\spacalgcomment{spacalgblue}{Fixed rules}
\ENDFOR
\STATE $\mathcal S\gets\textsc{Sort}[(m(\alpha),u)\mid(u,\alpha,G)\in\mathcal B]$\spacalgcomment{spacalgblue}{Morton order}
\FOR[\spacalgcomment{spacalgorange}{Exact recovery}]{each $(z,u)$ in $\mathcal S$}
    \STATE $\widehat{\mathcal M}\gets\widehat{\mathcal M}\cup\{(m^{-1}(z)+d,k)\mid(d,k)\in\mathcal P_u\}$
\ENDFOR
\RETURN $\mathcal M_{\mathrm{BPE}},\mathcal V,\mathcal P,\mathcal S,\widehat{\mathcal M}$
\end{algorithmic}
\end{wrapfloat}
\para{Local Context and the Unigram Gap.}
Let $\Omega\subset\mathbb{Z}^{3}$ be the set of occupied latent sites with leaf tokens $X_c\in\{0,\dots,v_0{-}1\}$. For $k\ge1$, define
\begin{align}
    \mathcal{N}_k(c) = \{c'\in\Omega:1\le\|c'-c\|_1\le k\}.
    \label{eq:block_neighborhood}
\end{align}
The optimal unigram predictor incurs expected loss $H(X_c)$, while the optimal local-context predictor incurs $H(X_c\mid X_{\mathcal{N}_k(c)})$. Their difference is the spatial mutual information
\begin{align}
\Delta_{\mathrm{uni}}
&= H(X_c)-H(X_c\mid X_{\mathcal{N}_k(c)})
\notag\\*
&= I(X_c;X_{\mathcal{N}_k(c)}),
\label{eq:unigram_gap}
\end{align}
termed the \emph{unigram gap}. A large gap indicates local information beyond marginal token frequencies, motivating merges of recurring spatial patterns into macro tokens.

\para{3D-BPE Macro Tokens.}
Let $\Delta^{+}=\{e_x,e_y,e_z\}$. Each current block stores a token ID, an anchor $\alpha$, and its occupied leaf coordinates $G$. Token $u$ has a relative leaf pattern $\mathcal P_u=\{(d,k)\}$, with $\mathcal P_k=\{(\mathbf 0,k)\}$ for a leaf token. Distinct blocks $A$ and $B$ are directed face-adjacent when $c\in G_A$ and $c+\delta\in G_B$ for some $\delta\in\Delta^{+}$. Candidate $p=(a,b,r)$ records the two token IDs and their anchor displacement $r=\alpha_B-\alpha_A$. Each resulting macro token represents a face-connected pattern of relative coordinates and leaf IDs. Each merge anchors the combined pattern at its lexicographically smallest leaf coordinate and updates all relative offsets.

\para{Tokenizer Training.}
Algorithm~\ref{alg:3d-bpe-train} first learns an ordered merge table and leaf patterns offline. The learned table is then fixed for new shapes. \textsc{InitBlocks} creates one block per leaf. \textsc{CountFacePairs} counts each directed block pair once per shape; \textsc{MatchPairs} orders matches lexicographically by $(\alpha_A,\alpha_B,a,b)$. Encoding starts at line~\ref{alg:bpe-encode-start}. \textsc{ApplyRule} constructs $Q$ and $s$ as in lines~\ref{alg:bpe-pattern-start} and~\ref{alg:bpe-anchor} using the supplied token ID, then repeats lines~\ref{alg:bpe-replace-start} to~\ref{alg:bpe-replace-end} on one shape. \textsc{Sort} orders coordinate and token pairs by increasing Morton coordinate, using the map $m$ in Eq.~\plaineqref{eq:morton}.


\para{Empirical Analysis.}
To characterize compression and macro-token boundary dependencies, we measure token and edge statistics. The directed leaf-edge multiset is
\begin{align}
    \mathcal{E}_{X}= \{(X_c,X_{c+\delta},\delta): c,c+\delta\in\Omega,\ \delta\in\Delta^{+}\}.
    \label{eq:leaf_edges}
\end{align}
Its empirical distribution $\hat{p}_X(a,b,\delta)$ gives the directed 3D adjacency entropy and mutual information
\begin{equation}
\begin{aligned}
\widehat{H}_{2}^{3D}(X)
  &= -\sum_{a,b,\delta}\hat{p}_X(a,b,\delta)\log_2\hat{p}_X(a,b,\delta),\\
\widehat{I}_{X}
  &= \widehat{I}\bigl((X_u,\delta);X_v\bigr)
   = \widehat{H}(X_u,\delta) + \widehat{H}(X_v) - \widehat{H}(X_u,X_v,\delta).
\end{aligned}
\label{eq:directed_edge_entropy}
\end{equation}
With block token IDs $B_u$, boundary contacts form
\[
\mathcal{E}_{B}^{\partial}
= \{(B_u,B_v,\delta): u\ne v,\ c\in G_u,\ c+\delta\in G_v,\ \delta\in\Delta^{+}\}.
\]
Counting each boundary leaf contact separately gives $\widehat{I}_{B}^{\partial}=\widehat{I}((B_u,\delta);B_v)$. The ratios are
\begin{equation}
\begin{aligned}
R_{\mathrm{MI}}
&= \frac{\widehat{I}_{B}^{\partial}}{\widehat{I}_{X}},
&
R_{\mathrm{edge}}
&= \frac{|\mathcal{E}_{B}^{\partial}|}{|\mathcal{E}_{X}|},
&
R_{\mathrm{len}}
&= \frac{\sum_i K_i}{\sum_i N_i},
\end{aligned}
\label{eq:bpe_metrics}
\end{equation}
where $N_i,K_i$ count sample $i$'s leaves and final blocks, including unmerged leaves. The macro-token unigram entropy $\widehat{H}_1(B)$ is defined analogously, and $R_{\mathrm{wMI}}=R_{\mathrm{MI}}R_{\mathrm{edge}}$ summarizes edge-weighted boundary dependence. Short-range sequence metrics $H_{\mathrm{bi}}(B), H_{\mathrm{tri}}(B)$ are the empirical conditional entropies of $B_t$ given the previous one/two tokens. Figure~\ref{fig:3d_bpe_sweep} and Table~\ref{tab:3d_bpe_baselines} report these quantities.

\para{Reversibility.}
The final loop in Algorithm~\ref{alg:3d-bpe-train} recovers each anchor from its Morton coordinate and expands $\mathcal P_u$ into leaves $(m^{-1}(z)+d,k)$. A leaf token uses its singleton pattern with zero offset. Decoding the encoded sequence exactly recovers the input sparse mapping $\mathcal M$.

\subsection{Morton Serialization and Unified Modeling}
\label{sec:morton_llm}

\firstpara{Injecting Spatial Inductive Bias via Morton Order.}
After 3D-BPE, each shape's macro-tokens are sorted by the Morton order of their anchors, yielding a deterministic LLM sequence. For an anchor $\mathbf{b}=(b_x,b_y,b_z)\in\{0,\dots,31\}^3$ with $L{=}5$ bits per axis,
\begin{equation}
m(\mathbf{b}) = \sum_{i=0}^{L-1}\Bigl[
  \bigl(\lfloor b_x/2^i\rfloor\bmod 2\bigr)2^{3i}
 +\bigl(\lfloor b_y/2^i\rfloor\bmod 2\bigr)2^{3i+1}
 +\bigl(\lfloor b_z/2^i\rfloor\bmod 2\bigr)2^{3i+2}
 \Bigr].
\label{eq:morton}
\end{equation}
Morton ordering promotes locality, while coordinates and 3D-BPE offsets retain explicit 3D structure.

\para{Sequence Format and Generation.}
The Sparse SDF VQ-VAE and 3D-BPE encode a shape as pairs of Morton coordinates and geometry tokens. The same LLM reads these pairs to produce captions and predicts them autoregressively for text-conditioned generation. As shown in the lower part of Fig.~\ref{fig:spac_dataflow}, predicted macro tokens are expanded into leaf tokens and block coordinates, the VQ-VAE decoder reconstructs the sparse SDF, and Marching Cubes extracts the mesh directly.

%% file: sec/experiments.tex
\section{Empirical Results} \label{sec:exp}

\subsection{Implementation Details}

\para{VQ-VAE Training.}
The Sparse SDF VQ-VAE uses SDF samples from $50$k ShapeNet+Objaverse meshes for geometry reconstruction and codebook learning. As detailed in Sec.~\ref{sec:sparse_mapping}, its first stage freezes the pretrained encoder--decoder to learn the codebook, and its second stage jointly fine-tunes the full VQ-VAE for sparse SDF reconstruction.

\para{LLM Training.}
The LLM uses text pairs for the same $50$k meshes, training on $8$ A100 GPUs with batch size $2$, learning rate $5{\times}10^{-5}$, and cosine decay. The Qwen3-4B decoder-only backbone adds $8192$ distance, $8192$ macro, and $32768$ Morton-coordinate tokens. Stage~1 freezes the VQ-VAE encoder and LLM backbone, training only new 3D-token embeddings and the output projection to align geometric symbols with the LLM's semantic space. Stage~2 keeps the encoder frozen and fully fine-tunes the LLM on text-paired 3D assets mixed with UltraChat for language retention, using AdamW, $10\%$ warmup, DeepSpeed ZeRO-2, and mixed precision for LLM optimization.

\par\smallskip
\noindent\begin{minipage}{\linewidth}
\centering
\captionsetup{skip=4pt,font=small}
\renewcommand{\arraystretch}{0.94}
\begingroup
\captionsetup{type=table}
\footnotesize
\caption{\textbf{Language capabilities comparison.}
Best values are bold and second-best values are underlined.}
\label{tab:language_capability}
\begin{tabular*}{\linewidth}{@{\extracolsep{\fill}}>{\raggedright\arraybackslash}p{0.54\linewidth}cccc@{}}
\toprule
\textbf{Model} & \textbf{MMLU} & \textbf{MMLU-Pro} & \textbf{GSM8K} & \textbf{BBH} \\
\midrule
SPAC\,\pubvenue{ours}
& \underline{69.5} & \underline{47.9} & \underline{86.1} & \textbf{73.4} \\
Qwen3-4B~\citep{yang2025qwen3}\,\pubvenue{arXiv'25}
& \textbf{72.9} & \textbf{50.5} & \textbf{87.7} & \underline{72.5} \\
ShapeLLM-Omni-7B~\citep{ye2025shapellmomni}\,\pubvenue{NeurIPS'25}
& 63.9 & 38.2 & 51.1 & 68.6 \\
Chameleon-7B~\citep{aghajanyan2024chameleon}\,\pubvenue{arXiv'24}
& 59.4 & 25.4 & 66.9 & 57.9 \\
LLaMA-Mesh-8B~\citep{wang2024llamamesh}\,\pubvenue{arXiv'24}
& 57.4 & 27.0 & 33.1 & 65.5 \\
PointLLM-13B~\citep{xu2023pointllm}\,\pubvenue{ECCV'24}
& 48.5 & 11.8 & 27.2 & 51.3 \\
\bottomrule
\end{tabular*}
\par\endgroup
\par\smallskip
\begingroup
\captionsetup{type=table}
\footnotesize
\caption{
\textbf{3D object captioning results on Objaverse.}
The table highlights the best result for each metric in bold and marks the second-best result with underlining.
}
\label{tab:3d_captioning}
\begin{tabular*}{\linewidth}{@{\extracolsep{\fill}}>{\raggedright\arraybackslash}p{0.43\linewidth}ccccc@{}}
\toprule
\textbf{Model}
& \textbf{BLEU-1}
& \textbf{ROUGE-L}
& \textbf{METEOR}
& \shortstack{\textbf{Sentence-}\\\textbf{BERT}}
& \textbf{SimCSE} \\
\midrule
\rowcolor[gray]{0.93}
\multicolumn{6}{l}{\textbf{\textit{General vision-language models}}} \\
\specialrule{\lightrulewidth}{0pt}{0pt}
InstructBLIP-13B~\citep{dai2023instructblip}\,\pubvenue{NeurIPS'23}
& 4.65 & 8.85 & 13.23 & 45.90 & 48.86 \\
LLaVA-13B~\citep{liu2023llava}\,\pubvenue{NeurIPS'23}
& 4.02 & 8.15 & 12.58 & 46.37 & 45.90 \\
Qwen2.5-VL-7B~\citep{bai2025qwen25vl}\,\pubvenue{arXiv'25} & 4.05 & 7.85 & 14.23 & 48.90 & \underline{50.86} \\
\midrule
\rowcolor[gray]{0.93}
\multicolumn{6}{l}{\textbf{\textit{3D understanding models}}} \\
\specialrule{\lightrulewidth}{0pt}{0pt}
3D-LLM~\citep{hong2023injecting3d}\,\pubvenue{NeurIPS'23}
& 16.91 & 19.48 & 19.73 & 44.48 & 43.68 \\
PointLLM-13B~\citep{xu2023pointllm}\,\pubvenue{ECCV'24}
& 17.09 & 20.99 & 16.45
& \underline{50.15} & 50.83 \\
LEO~\citep{huang2024leo}\,\pubvenue{ICML'24} & 16.98 & 20.12 & \underline{20.91} & 48.01 & 47.25 \\
\midrule
\rowcolor[gray]{0.93}
\multicolumn{6}{l}{\textbf{\textit{Unified 3D understanding and generation models}}} \\
\specialrule{\lightrulewidth}{0pt}{0pt}
ShapeLLM-Omni~\citep{ye2025shapellmomni}\,\pubvenue{NeurIPS'25}
& \underline{18.51} & \underline{21.37} & 19.89
& 48.34 & 49.72 \\
SPAC (ours)
& \textbf{22.31} & \textbf{24.43} & \textbf{21.50} & \textbf{68.24} & \textbf{62.20} \\
\bottomrule
\end{tabular*}
\par\endgroup
\par\smallskip
\begingroup
\captionsetup{type=table}
\footnotesize
\setlength{\heavyrulewidth}{0.6pt}
\setlength{\lightrulewidth}{0.4pt}
\caption{
\textbf{Comparison on text-to-3D generation.}
KD is scaled by $10^2$. Best are in bold and second-best are underlined.
}
\label{tab:text_image_to_3d}
\begin{tabular*}{\linewidth}{@{\extracolsep{\fill}}>{\raggedright\arraybackslash}p{0.57\linewidth}ccc@{}}
\toprule
\textbf{Method}
& \textbf{CLIP$\uparrow$}
& \textbf{FD$_{\mathrm{incep}}\downarrow$}
& \textbf{KD$_{\mathrm{incep}}\downarrow$} \\
\midrule
\rowcolor[gray]{0.93}
\multicolumn{4}{l}{\textbf{\textit{Specialized 3D generators}}} \\
\specialrule{\lightrulewidth}{0pt}{0pt}
SAR3D~\citep{chen2025sar3d}\,\pubvenue{CVPR'25}
& 23.9 & 27.2 & 0.28 \\
3DTopia-XL~\citep{chen20253dtopia}\,\pubvenue{CVPR'25} & 25.89 & 43.46 & 1.18 \\
Shap-E~\citep{jun2023shape}\,\pubvenue{arXiv'23} & 24.94 & 53.24 & 1.13 \\
CLAY~\citep{zhang2024clay}\,\pubvenue{TOG'24} & 23.28 & 48.43 & 0.59 \\
TRELLIS~\citep{xiang2025structured}\,\pubvenue{CVPR'25}
& \textbf{30.8} & \textbf{18.3} & \textbf{0.10} \\
\midrule
\rowcolor[gray]{0.93}
\multicolumn{4}{l}{\textbf{\textit{Unified 3D models}}} \\
\specialrule{\lightrulewidth}{0pt}{0pt}
ShapeLLM-Omni~\citep{ye2025shapellmomni}\,\pubvenue{NeurIPS'25}
& \underline{26.7} & 25.9 & 0.25 \\
SPAC (ours)
& 24.4 & \underline{23.7} & \underline{0.22} \\
\bottomrule
\end{tabular*}
\par\endgroup

\end{minipage}\par

\subsection{Comparison with the State-of-the-Art}

\firstpara{Results on Language and Conversational Abilities.}
Table~\ref{tab:language_capability} compares SPAC with Qwen3-4B~\citep{yang2025qwen3} and the 3D-capable baselines PointLLM~\citep{xu2023pointllm}, LLaMA-Mesh~\citep{wang2024llamamesh}, Chameleon~\citep{aghajanyan2024chameleon}, and ShapeLLM-Omni~\citep{ye2025shapellmomni} on MMLU~\citep{hendrycks2020measuring}, MMLU-Pro~\citep{wang2024mmluprorobustchallengingmultitask}, GSM8K~\citep{cobbe2021training}, and BBH~\citep{suzgun2022challengingbigbenchtaskschainofthought}. After joint fine-tuning, SPAC leads the compared 3D-oriented LLMs on all four benchmarks. It scores $69.5$, $47.9$, and $86.1$ on MMLU, MMLU-Pro, and GSM8K, respectively, and raises BBH from $72.5$ to $73.4$ relative to the Qwen3-4B backbone.

\firstpara{Results on 3D-to-Text Captioning.}
Following PointLLM~\citep{xu2023pointllm}, we use its curated test set and evaluation metrics. SPAC leads all five metrics in Table~\ref{tab:3d_captioning}, with METEOR of 21.50 versus 20.91 for LEO and SimCSE of 62.20 versus 50.86 for the strongest baseline. Gains span lexical-overlap and sentence-embedding metrics, supporting both reference agreement and semantic relevance.

\firstpara{Results on Text-to-3D Generation.}
In Table~\ref{tab:text_image_to_3d}, SPAC achieves FD of 23.7 and KD of 0.22, compared with 25.9 and 0.25 for ShapeLLM-Omni. Its CLIP score of 24.4 complements these distributional metrics with an assessment of alignment between generated shapes and conditioning text. Following ShapeLLM-Omni, we use top-$p=0.7$ and temperature $0.7$ for generation.

\par\medskip
\noindent\begin{minipage}{\linewidth}
\centering
\captionsetup{skip=5pt,font=small,hypcap=false}
\includegraphics[width=\linewidth]{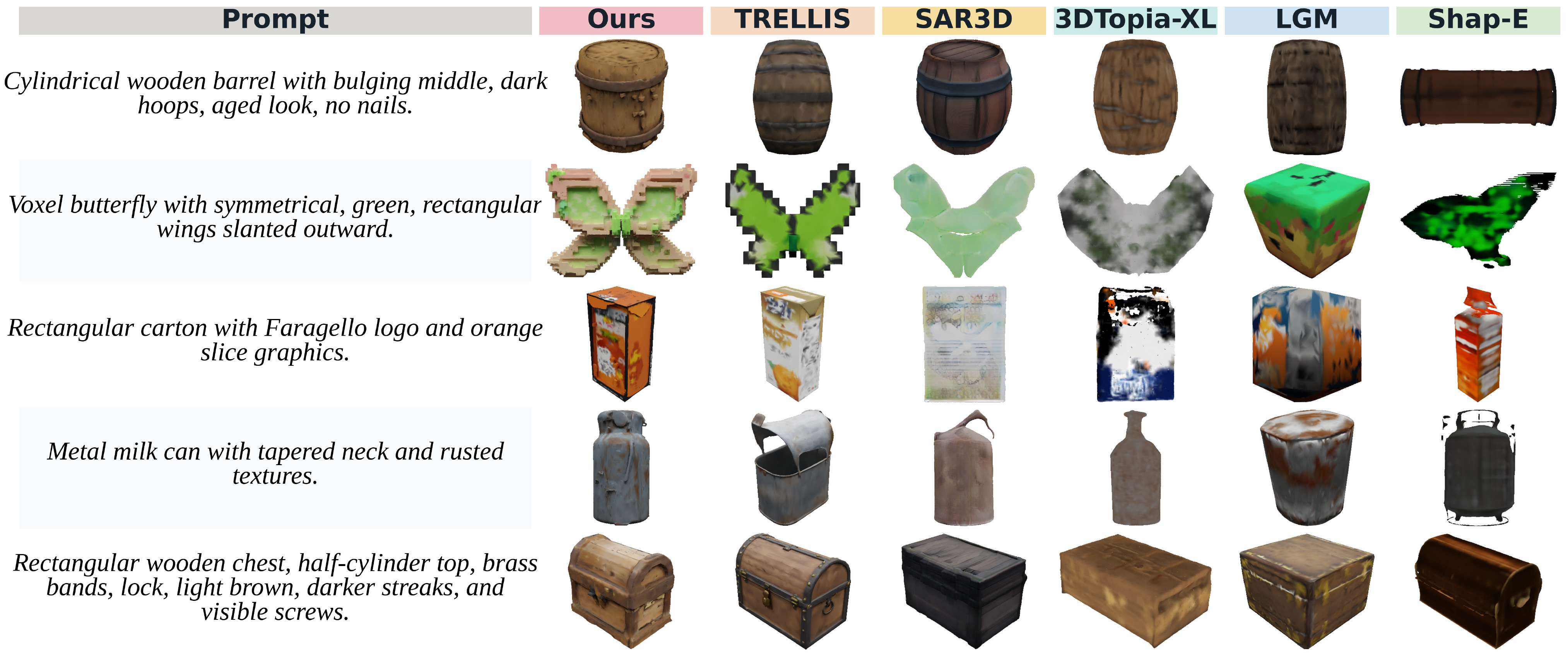}
\captionof{figure}{\textbf{Comparisons with other baselines on the Text-to-3D task.} SPAC generates untextured meshes. Following ShapeLLM-Omni~\citep{ye2025shapellmomni}, we use TRELLIS~\citep{xiang2025structured} to add surface appearance to the SPAC results shown here.}
\label{fig:generation_case_study}
\end{minipage}\par

\subsection{Ablation and Analysis}
\label{sec:efficiency}

\firstpara{3D-BPE.}
In Table~\ref{tab:spac_ablation}, SPAC improves generation and all five captioning metrics over both ablations while reducing FLOPs and latency. Relative to the Morton variant, METEOR increases from 15.59 to 21.50 and SimCSE from 58.04 to 62.20. At equal merge budgets (Table~\ref{tab:3d_bpe_baselines}), 3D-BPE's sequence-length ratio is $0.754$, below Morton 1D ($0.803$), Hilbert 1D ($0.820$), and Random 1D ($0.998$). Its lowest length-normalized unigram cost and short-range dependency measures support face adjacency over 1D proximity. On 43,180 text-to-3D records, 8,192 rules reduce full reference lengths (including coordinates, text, and control tokens) from 6,873.2 to 3,753.1 tokens on average (45.4\%) and from $14{,}675.1$ to $8{,}483.2$ at P95 ($42.2\%$).

\par\smallskip
\noindent\begin{minipage}{\linewidth}
\centering
\captionsetup{skip=5pt,font=small}
\begingroup
\captionsetup{type=table}
\footnotesize
\setlength{\tabcolsep}{3.5pt}
\setlength{\heavyrulewidth}{0.6pt}
\setlength{\lightrulewidth}{0.4pt}
\caption{\textbf{3D-BPE ablation.} KD is scaled by $10^2$. The best result in each column is bold, and the second best is underlined.}
\label{tab:spac_ablation}
\begin{adjustbox}{max width=\linewidth}\begin{tabular}{@{}l|ccc|ccccc|cc@{}}
\toprule
& \multicolumn{3}{c}{\textbf{Text-to-3D}}
& \multicolumn{5}{c}{\textbf{Captioning}}
& \multicolumn{2}{c}{\textbf{Cost}} \\
\cmidrule(lr){2-4}\cmidrule(lr){5-9}\cmidrule(l){10-11}
\textbf{Variant} & \textbf{CLIP}$\uparrow$ & \textbf{FD}$\downarrow$ & \textbf{KD}$\downarrow$
& \textbf{BLEU-1}$\uparrow$ & \textbf{ROUGE-L}$\uparrow$ & \textbf{METEOR}$\uparrow$
& \shortstack{\textbf{Sentence-}\\\textbf{BERT}$\uparrow$} & \textbf{SimCSE}$\uparrow$
& \shortstack{\textbf{FLOPs}\\\textbf{G}$\downarrow$} & \shortstack{\textbf{Latency}\\\textbf{s}$\downarrow$} \\
\midrule
w.o. BPE & 22.1 & 28.4 & 0.31 & 7.08 & 19.72 & 13.07 & 56.27 & 51.09 & 69,771.35 & 348.40 \\
Morton 1D-BPE & \underline{23.4} & \underline{26.9} & \underline{0.27} & \underline{9.15} & \underline{21.72} & \underline{15.59} & \underline{62.61} & \underline{58.04} & \underline{56,299.82} & \underline{262.93} \\
SPAC & \textbf{24.4} & \textbf{23.7} & \textbf{0.22} & \textbf{22.31} & \textbf{24.43} & \textbf{21.50} & \textbf{68.24} & \textbf{62.20} & \textbf{41,122.05} & \textbf{171.69} \\
\bottomrule
\end{tabular}\end{adjustbox}
\par\endgroup
\par\medskip
\begingroup
\captionsetup{type=table}
\caption{\textbf{BPE compression and full LLM reference sequence lengths.} Results use 8,192 merge rules. $R_{\mathrm{len}}$ is the retained geometry-token fraction; entropy metrics are in bits. Full sequence lengths count geometry, coordinates, text, and control tokens before truncation or packing; reduction is relative to Leaf. Best are in bold and second-best are underlined.}
\label{tab:3d_bpe_baselines}
\small
\setlength{\heavyrulewidth}{0.6pt}
\setlength{\lightrulewidth}{0.4pt}
\begin{tabular*}{\linewidth}{@{\extracolsep{\fill}}l|cccc|ccc@{}}
\toprule
& \multicolumn{4}{c}{\textbf{Tokenizer analysis}}
& \multicolumn{3}{c}{\textbf{Text-to-3D lengths}}\\
\cmidrule(lr){2-5}\cmidrule(l){6-8}
\textbf{Method} & $R_{\mathrm{len}}\downarrow$ & $\widehat{H}_1R_{\mathrm{len}}\downarrow$ &
$H_{\mathrm{bi}}\downarrow$ & $H_{\mathrm{tri}}\downarrow$ &
\textbf{Mean}$\downarrow$ & \textbf{P95}$\downarrow$ & \textbf{Reduction}$\uparrow$\\
\midrule
Leaf & 1.000 & 10.910 & \textemdash & \textemdash & 6,873.2 & 14,675.1 & \textemdash\\
Morton 1D & \underline{0.803} & \underline{9.881} & \underline{1.092} & 0.127 & 5,443.2 & 11,852.2 & 20.81\%\\
Hilbert 1D & 0.820 & 10.010 & 1.220 & \underline{0.072} & \underline{5,385.4} & \underline{11,700.1} & \underline{21.65\%}\\
Random 1D & 0.998 & 10.910 & 2.167 & 0.166 & 5,645.6 & 12,281.1 & 17.86\%\\
3D-BPE & \textbf{0.754} & \textbf{9.420} & \textbf{0.972} & \textbf{0.061} &
\textbf{3,753.1} & \textbf{8,483.2} & \textbf{45.40\%}\\
\bottomrule
\end{tabular*}
\par\endgroup
\par\medskip
\captionsetup{type=figure}
\includegraphics[width=\linewidth]{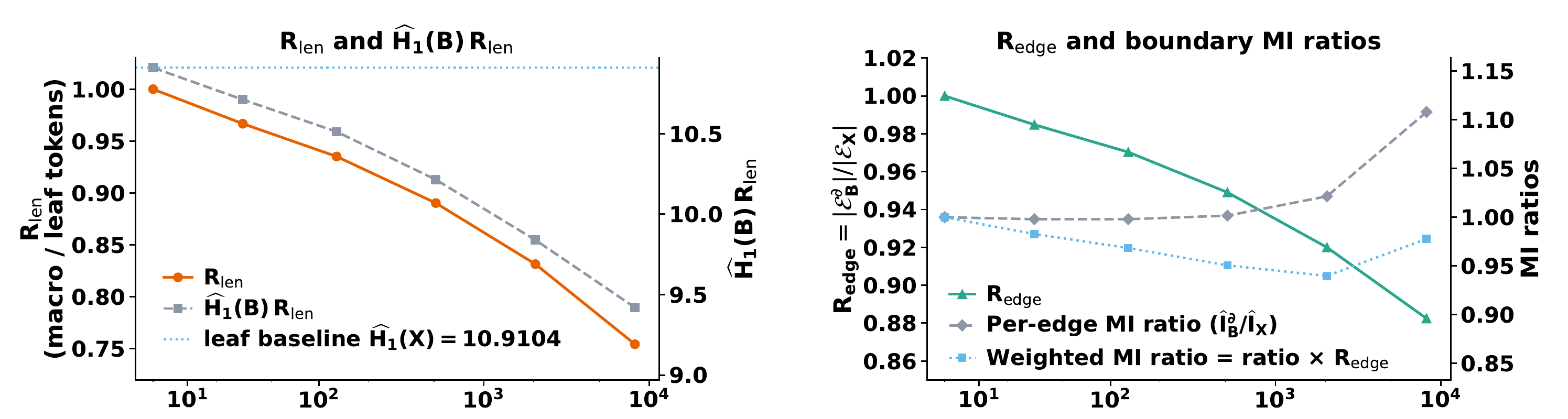}
\caption{
\textbf{3D-BPE merge budget sweep.}
Horizontal axis: merge budget $K_{\mathrm{merge}}$. Left: length compression ratio $R_{\mathrm{len}}$ and length-normalized unigram cost $\widehat{H}_1(B)R_{\mathrm{len}}$. The dotted line marks the leaf baseline $\widehat{H}_1(X)$. Right: residual boundary edge ratio $R_{\mathrm{edge}}$, per-edge MI ratio $R_{\mathrm{MI}}$, and weighted MI ratio $R_{\mathrm{wMI}}$. Ratios are relative to unmerged leaf tokens.
}
\label{fig:3d_bpe_sweep}

\end{minipage}\par

\para{Downstream Quality and Cost.}
Relative to Morton 1D-BPE, SPAC improves CLIP from $23.4$ to $24.4$ and FD from $26.9$ to $23.7$, while reducing FLOPs by $27.0\%$ and latency by $34.7\%$. The comparison measures the combined effect of spatial grouping and its resulting sequence length.

\para{Interpreting the Merge-Budget Sweep.}
Merging removes internal contacts from the boundary statistics in Fig.~\ref{fig:3d_bpe_sweep}. Higher per-edge MI can therefore coexist with fewer boundary contacts and lower edge-weighted dependence. The decreasing token count and length-normalized unigram cost jointly characterize compression.

\Needspace{270pt}
\begin{wrapfigure}[20]{r}{0.42\linewidth}
\centering
\captionsetup{font=footnotesize,justification=raggedright,skip=4pt}
\includegraphics[width=\linewidth]{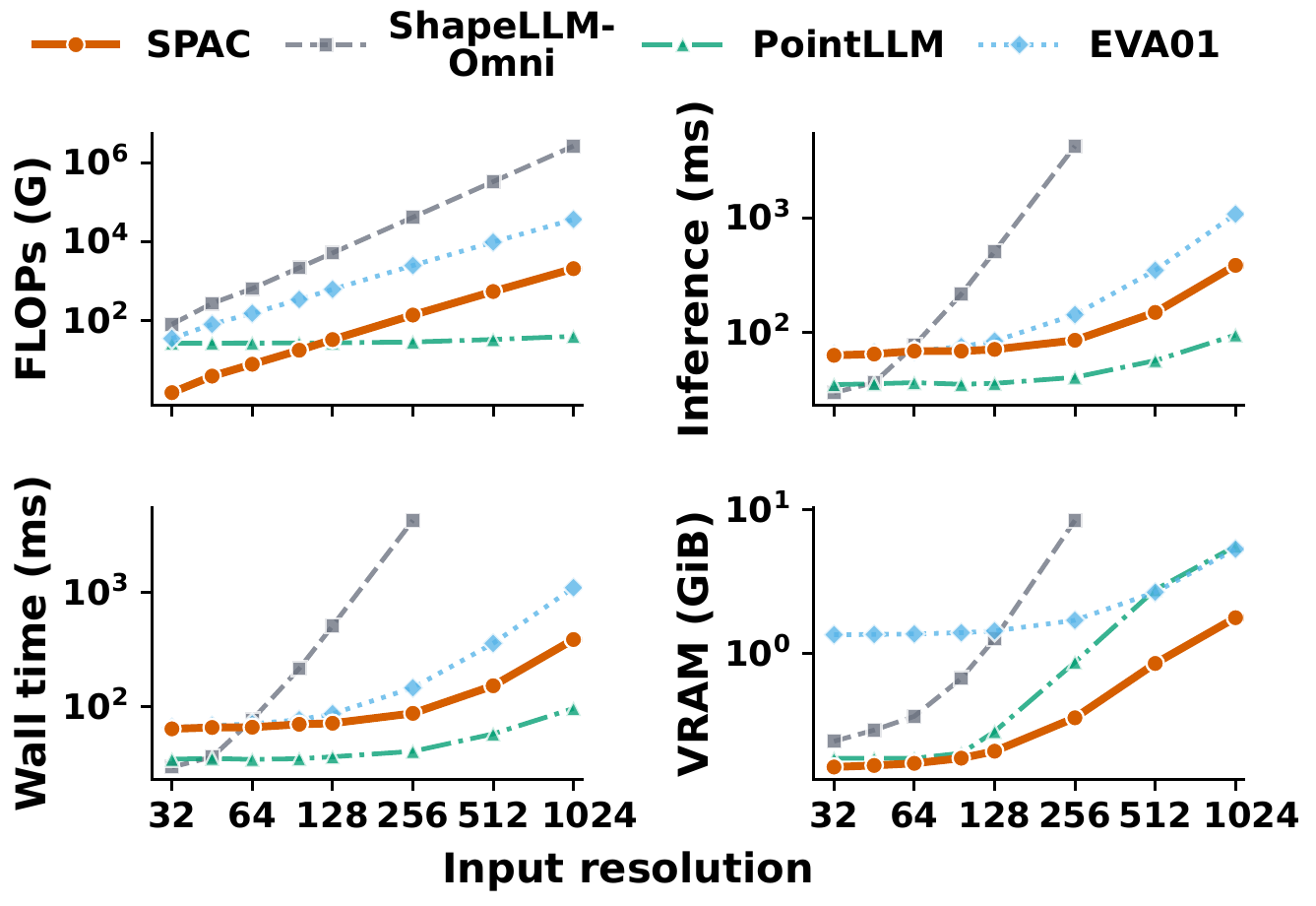}
\caption{
\textbf{Encoder efficiency across input resolutions.}
The top-left and top-right panels report forward FLOPs and encoder inference latency; the bottom-left and bottom-right panels report prepared-input wall time and peak inference VRAM. Results use batch size $1$ and logarithmic vertical axes. \textsc{SPAC}'s sparse encoder exhibits near-$O(R^2)$ scaling with low memory growth. PointLLM's fixed $512$-group budget yields nearly constant FLOPs, while EVA01 incurs a larger constant cost. Dense FLOPs beyond the executable range are analytically extrapolated.
}
\label{fig:cmp}
\end{wrapfigure}
\para{Encoder Efficiency.}
We compare \textsc{SPAC} with ShapeLLM-Omni's dense 3D VQ-VAE~\citep{ye2025shapellmomni}, PointLLM's Point-BERT~\citep{xu2023pointllm}, and EVA01's generation-side O-Voxel sparse VAE~\citep{yang2026eva01unifiednative3d}, omitting EVA01's understanding-side Point-BERT due to its overlap with PointLLM. Figure~\ref{fig:cmp} reports FLOPs, latency, prepared-input time, and peak memory across resolutions. By processing a narrow surface band, \textsc{SPAC} scales better than ShapeLLM-Omni: at the largest uncapped resolution, $R=256$, it requires $137.98$ G FLOPs and $85.04$ ms, versus ShapeLLM-Omni's $41.38$ T FLOPs and $4227.84$ ms. ShapeLLM-Omni has lower latency at low $R$, followed by cubic cost growth with resolution. PointLLM's $512$-group budget yields nearly constant FLOPs and lower latency, with sampling and neighborhood intermediates increasing memory with resolution. EVA01 combines surface sparsity with a larger encoder and a higher constant cost.

Encoder costs and LLM sequence lengths measure distinct stages. Comparisons of native encoder configurations jointly reflect architecture, representation, and sparsity, characterizing the full encoder under its own settings.

\begin{samepage}
\para{Complexity Analysis.}
For a fixed regular normalized surface of area $A$, band volume is $O(A\tau)$. Set $h=1/R$ and $\tau=\eta h$ with fixed voxel width $\eta$. For fixed architecture and local-attention neighborhoods, $N_\ell=O(N_{\mathrm{active}})$, giving
\begin{gather}
N_{\mathrm{active}}(R)
=O\!\left(\frac{A\tau}{h^3}\right)
=O(A\eta R^2),
\label{eq:active_voxel_complexity}\\*
\sum\nolimits_\ell O\!\left(N_\ell k_\ell^3 C_{\ell,\mathrm{in}}C_{\ell,\mathrm{out}}\right)
+\sum\nolimits_\ell O(N_\ell)=O(R^2).
\label{eq:encoder_complexity}
\end{gather}
\end{samepage}
At fixed $A$ and batch size, active sites and encoder cost therefore scale as $O(R^2)$, compared with $O(R^3)$ for dense encoding. Fitted active-input and FLOP slopes are $1.991$ and $2.164$; deviations reflect finite-resolution boundary and downsampling effects. Surface-sampling-capped samples are excluded. A fixed physical-width band instead activates $O(R^3)$ voxels. LLM attention requires $O(L^2)$ interactions for sequence length $L$. The appendix details the evaluation protocol and the associated metrics.

%% file: sec/conclusion.tex
\section{Conclusion} \label{sec:conc}

SPAC uses a shared sparse SDF vocabulary learned by a Sparse SDF VQ-VAE to unify 3D understanding and generation in a single autoregressive LLM. Reversible 3D-BPE compresses recurring spatial patterns, and the predicted tokens decode directly into mesh geometry. Experiments show favorable sparse encoder scaling, shorter complete LLM sequences, and gains in both captioning and generation over leaf tokens and Morton 1D-BPE. Exact expansion preserves the quantized coordinate--token mapping; SDF sampling, the codebook, and the decoder determine surface detail. SPAC models geometry and text, generating untextured meshes from text prompts. Future work includes image conditioning and surface appearance, broader training coverage across shapes and descriptions, and refined tokenization for thin structures and less common shapes.

\section*{LLM Usage Statement}

We used OpenAI ChatGPT and Codex to assist with manuscript revision and
language polishing, figure preparation and revision, and code implementation,
debugging, and maintenance. The authors directed this assistance and retain
full responsibility for the manuscript, figures, code, experimental results,
and conclusions.

%% file: sec/appendix.tex
\clearpage
\appendix
\phantomsection
\addcontentsline{toc}{section}{Supplementary Material}
\begingroup
\centering
\Large
\parbox{0.78\linewidth}{\centering\bfseries\papertitle}\\[0.2em]
Supplementary Material\par
\vspace{0.5em}
\endgroup
\setcounter{figure}{0}
\renewcommand{\thefigure}{A\arabic{figure}}
\renewcommand{\theHfigure}{appendix.\arabic{figure}}

We detail the training and inference pipeline in Sec.~\ref{app:training} and the implementation settings in Sec.~\ref{app:implementation}. Section~\ref{app:3dbpe_details} develops the 3D-BPE algorithm, its properties, and a controlled synthetic validation. Sections~\ref{app:compression} and~\ref{app:efficiency} analyze compression and computational cost, including an analytic-SDF grid experiment. Sections~\ref{sec:limit} and~\ref{sec:impacts} discuss limitations and societal impact. Section~\ref{app:qualitative} concludes with additional qualitative results for both generation and understanding.

\section{Training and Inference Details}
\label{app:training}

\subsection{From Geometry to Sparse Tokens}
\label{app:training_geometry}

Figure~\ref{fig:sparc} maps a normalized mesh to sparse latent coordinates and geometry codes. The encoder downsamples the $R=256$ SDF grid by $8\times$ to a $32^3$ latent lattice with $C_z=16$ channels. SDF coordinates $\mathbf{x}$ and latent coordinates $c_i$ therefore index different grids.

For the signed distance $s(\mathbf{x})$, the supervision band has physical width $\tau_w=\eta_w/R$, with $\eta_w=4.0$. Its normalized values are
\begin{equation}
\bar{s}(\mathbf{x})
=\operatorname{clip}\!\left(s(\mathbf{x})/\tau_w,-1,1\right).
\label{eq:app_sdf_normalization}
\end{equation}
The supervision band satisfies $|s(\mathbf{x})|<\tau_w$. The encoder uses its inner subset $|\bar{s}(\mathbf{x})|\leq\eta_e/\eta_w$ with $\eta_e=0.5$, within $\eta_e/R$ of the surface. This concentrates computation near the surface, while the wider band provides reconstruction targets and sparse support.

The sparse 3D U-Net adapted from Direct3D-S2~\citep{wu2025direct3ds2} maps these inputs to active latent features $\mathbf{z}_i$. The codebook $\mathbf{E}\in\mathbb{R}^{K\times C_z}$ contains $K=8192$ entries. Nearest-neighbor quantization and its straight-through expression are
\begin{align}
k_i^*&=\arg\min_{k\in[K]}\|\mathbf{z}_i-\mathbf{e}_k\|_2,
\notag\\
\hat{\mathbf{z}}_i
&=\mathbf{z}_i+\mathrm{sg}(\mathbf{e}_{k_i^*}-\mathbf{z}_i).
\label{eq:app_st_quantization}
\end{align}
The forward value $\hat{\mathbf{z}}_i$ is the selected codebook vector. The stop-gradient operator $\mathrm{sg}$ makes its derivative with respect to $\mathbf{z}_i$ the identity, passing reconstruction gradients to the encoder. Leaf IDs and latent coordinates form the 3D-BPE input $\mathcal{M}=\{(c_i,k_i)\}_{i=1}^{M}$ in Eq.~\eqref{eq:sparse_map}.

\subsection{Two-Stage VQ-VAE Training}
\label{app:training_vq}

Stage one freezes the pretrained encoder and decoder and learns the codebook with exponential moving averages and K-means reinitialization. Stage two fine-tunes the encoder--decoder with reconstruction and commitment losses while continuing EMA codebook updates.

Reconstruction supervision uses three groups: regular surface samples, sharp geometric features, and contextual samples in the wider band. Let $\mathcal{G}=\{\mathrm{surf},\mathrm{feat},\mathrm{ctx}\}$ and let $\Omega_r$ contain the sampled target locations in group $r$. For nonempty groups, define
\begin{align}
\mathcal{L}_r
&=\frac{1}{|\Omega_r|}\sum_{\mathbf{x}\in\Omega_r}
\bigl(\bar{s}(\mathbf{x})-\tilde{s}(\mathbf{x})\bigr)^2,
\notag\\
\mathcal{L}_{\mathrm{recon}}
&=\sum_{r\in\mathcal{G}}\lambda_r\mathcal{L}_r.
\label{eq:app_group_reconstruction}
\end{align}
Here $\tilde{s}$ is the reconstructed normalized SDF and $\lambda_{\mathrm{surf}}=\lambda_{\mathrm{feat}}=\lambda_{\mathrm{ctx}}=1$. Averaging within each group preserves separate contributions from surface, feature, and contextual samples regardless of their counts.

The commitment term averages over all $M$ active latent sites in the current shape:
\begin{align}
\mathcal{L}_{\mathrm{commit}}
&=\frac{1}{M}\sum_{i=1}^{M}
\|\mathbf{z}_i-\mathrm{sg}(\mathbf{e}_{k_i^*})\|_2^2,
\notag\\
\mathcal{L}
&=\mathcal{L}_{\mathrm{recon}}+0.25\mathcal{L}_{\mathrm{commit}}.
\label{eq:app_vq_objective}
\end{align}
Commitment pulls encoder features toward fixed codebook targets. The wider band defines valid sparse support at each decoder scale, with reconstruction errors inside this support weighted by the three groups.

\subsection{Offline Compression and Two-Stage Autoregressive LLM Training}
\label{app:training_llm}

3D-BPE learns an ordered table of macro tokens and their leaf IDs and relative coordinates, then fixes it for encoding and expansion. Figure~\ref{fig:spac_dataflow} illustrates this shared table, and Algorithm~\ref{alg:3d-bpe-train} gives all three stages.

Qwen3-4B~\citep{yang2025qwen3} adds $8192$ distance tokens, $8192$ macro tokens, and $32768$ Morton-coordinate tokens to its vocabulary. Macro tokens identify learned patterns and Morton tokens locate anchors on the $32^3$ lattice, jointly modeled with text by the decoder-only LLM.

Alignment freezes the VQ-VAE encoder and LLM backbone, training only new 3D embeddings and the output projection. Joint fine-tuning keeps the encoder frozen and trains the full LLM on paired 3D assets and UltraChat. The encoder and merge table remain fixed.

Both geometric task directions use next-token prediction. For a conditioning prefix $x$ and a target continuation $y=(y_1,\ldots,y_T)$, the conditional likelihood factorizes as
\begin{equation}
p_\theta(y\mid x)
=\prod_{t=1}^{T}p_\theta(y_t\mid x,y_{<t}).
\label{eq:app_conditional_ar}
\end{equation}
Captioning predicts a textual continuation from geometry, whereas generation predicts a geometric continuation from text. UltraChat contributes conversational text during joint fine-tuning to support language retention.

\subsection{Bidirectional Inference}
\label{app:training_inference}

For captioning, the fixed VQ-VAE encodes the mesh SDF. Fixed 3D-BPE rules produce coordinate--geometry pairs, ordered by Morton anchor keys for the LLM to read. Coordinates locate patterns and Morton keys determine the serialized sequence order used by the LLM.

For generation, the LLM predicts coordinate--geometry pairs from text with top-$p=0.7$ and temperature $0.7$. Inverse Morton coding recovers anchors, and pattern lookup supplies leaf IDs and relative offsets. Adding anchors restores latent coordinates, and codebook lookup supplies features for VQ-VAE decoding into sparse SDF and Marching Cubes mesh extraction, as shown in Fig.~\ref{fig:spac_dataflow}.

\Needspace{6\baselineskip}
\section{Implementation Details}
\label{app:implementation}

\subsection{Model and Training Configuration}
\label{app:implementation_configuration}

Table~\ref{tab:app_training_settings} summarizes the configuration. The LLM keeps the trained VQ-VAE encoder frozen, and the main compression comparison uses $8192$ merge rules.

\begin{table}[!htb]
\centering
\small
\setlength{\tabcolsep}{3pt}
\caption{Model, training, and sampling settings. LLM optimization settings apply to language-model training; the tokenizer stages are described in Sec.~\ref{app:training_vq}.}
\label{tab:app_training_settings}
\begin{tabularx}{\linewidth}{@{}>{\raggedright\arraybackslash}p{0.38\linewidth}>{\raggedright\arraybackslash}X@{}}
\toprule
\textbf{Component} & \textbf{Setting} \\
\midrule
\multicolumn{2}{@{}l}{\textit{Geometry tokenizer}} \\
Training geometry & $50$k ShapeNet+Objaverse meshes \\
SDF / latent resolution & $256^3$ / $32^3$ \\
Downsampling / channels & $8\times$ / $C_z=16$ \\
Band widths & $\eta_w=4.0$, $\eta_e=0.5$ \\
Reconstruction weights & $\lambda_{\mathrm{surf}}=\lambda_{\mathrm{feat}}=\lambda_{\mathrm{ctx}}=1$ \\
Commitment weight & $\lambda_c=0.25$ \\
Leaf codebook size & $8192$ entries \\
\midrule
\multicolumn{2}{@{}l}{\textit{Spatial compression}} \\
Main merge budget & $8192$ rules \\
Serialization & Morton order of anchors \\
\midrule
\multicolumn{2}{@{}l}{\textit{Language model}} \\
Backbone & Qwen3-4B \\
Added vocabulary & $8192$ distance, $8192$ macro, $32768$ Morton-coordinate tokens \\
3D training data & Text pairs for the same $50$k meshes \\
Language data, stage 2 & UltraChat \\
Hardware / batch size & $8$ A100 GPUs / $2$ \\
Learning rate / schedule & $5\times10^{-5}$ / cosine decay \\
Stage 2 optimizer & AdamW; $10\%$ warmup \\
Stage 2 implementation & ZeRO-2; mixed precision \\
\midrule
\multicolumn{2}{@{}l}{\textit{Text-to-3D sampling}} \\
Nucleus sampling & top-$p=0.7$ \\
Temperature & $0.7$ \\
\bottomrule
\end{tabularx}
\end{table}

\subsection{Data and Task Construction}
\label{app:implementation_data}

SDF samples from $50$k ShapeNet and Objaverse meshes train reconstruction and codebook learning. Text pairs for the same meshes support both captioning from geometry and geometry prediction from text. Joint LLM fine-tuning also includes UltraChat conversations.

\subsection{Evaluation Settings}
\label{app:implementation_evaluation}

\para{Captioning.}
We follow the curated test set and evaluation scheme of PointLLM~\citep{xu2023pointllm}. SPAC receives SDF geometry and generates captions. Table~\ref{tab:3d_captioning} reports BLEU-1, ROUGE-L, METEOR, Sentence-BERT, and SimCSE; higher scores are better for all five metrics.

\para{Text-to-3D.}
Generation uses top-$p=0.7$ and temperature $0.7$, following ShapeLLM-Omni~\citep{ye2025shapellmomni}. Table~\ref{tab:text_image_to_3d} reports CLIP~\citep{radford2021learning}, $\mathrm{FD}_{\mathrm{incep}}$, and $\mathrm{KD}_{\mathrm{incep}}$. Higher CLIP and lower FD and KD are better. KD is displayed with the $10^2$ scaling used in the main comparison and ablation tables.

\para{Language Benchmarks.}
We evaluate language capabilities on MMLU~\citep{hendrycks2020measuring}, MMLU-Pro~\citep{wang2024mmluprorobustchallengingmultitask}, GSM8K~\citep{cobbe2021training}, and BBH~\citep{suzgun2022challengingbigbenchtaskschainofthought}. Table~\ref{tab:language_capability} compares the jointly fine-tuned SPAC model with its Qwen3-4B backbone and the 3D-capable language models.

\subsection{Schematic Task Sequences}
\label{app:implementation_sequences}

Let $T$ denote text and let the geometric payload be
\begin{equation}
\mathcal{S}
=\bigl((m(\alpha_1),u_1),\ldots,(m(\alpha_J),u_J)\bigr),
\label{eq:app_geometric_payload}
\end{equation}
where $J$ is the number of surviving blocks, $u_j$ is a leaf or macro token ID, and the anchor keys are in increasing Morton order. Each pair places its coordinate before its geometry token, following Algorithm~\ref{alg:3d-bpe-train}. The schematic task sequences are
\begin{equation}
\begin{aligned}
\text{3D-to-text:}\quad &\mathcal{S}\ \longrightarrow\ T,\\
\text{Text-to-3D:}\quad &T\ \longrightarrow\ \mathcal{S}.
\end{aligned}
\label{eq:app_task_directions}
\end{equation}
Arrows separate conditioning from prediction, and parentheses group coordinate--geometry pairs. Leaves expand to one code at zero offset and macros to learned patterns, connecting both tasks to the same sparse mapping and decoder.

\section{Detailed 3D-BPE Algorithm and Properties}
\label{app:3dbpe_details}

\subsection{Spatial Objects and Notation}
\label{app:3dbpe_notation}

The input to 3D-BPE is the quantized mapping
$\mathcal M=\{(c_i,k_i)\}_{i=1}^{M}$ from
Eq.~\ref{eq:sparse_map}, with distinct latent coordinates
$c_i\in\Omega\subseteq\{0,\ldots,31\}^3$.
These coordinates index the $32^3$ latent lattice produced from the
$256^3$ SDF input. We use the zero-based leaf IDs
$k_i\in\{0,\ldots,v_0-1\}$ of Algorithm~\ref{alg:3d-bpe-train};
this is a fixed relabeling of the codebook indices in Eq.~\ref{eq:vq}.
The leaf vocabulary has $v_0=8192$ entries. The merge budget $m$
bounds the number of added token types, so that
$|\mathcal V|-v_0\leq m$. For object $i$, $N_i$ is its input leaf
count and $K_i$ its number of final blocks, including unmerged leaves.
Thus, vocabulary size, merge budget, and per-object sequence length
refer to different quantities.

A current block is $A=(a,\alpha_A,G_A)$, comprising its token ID,
anchor, and support of occupied leaf coordinates. The pattern table
stores $\mathcal P_a=\{(d,k)\}$, where $d\in\mathbb Z^3$ is an offset
from the anchor and $k$ is a leaf ID. Its expansion is
\begin{equation}
\operatorname{Expand}(A)
=\{(\alpha_A+d,k):(d,k)\in\mathcal P_a\}.
\label{eq:app_block_expansion}
\end{equation}
A leaf uses $\mathcal P_k=\{(\mathbf 0,k)\}$.
The anchor is the lexicographically smallest coordinate in $G_A$.
Lexicographic minimization selects an occupied coordinate; it does
not take independent minima along the three axes. Consequently,
a relative pattern can contain negative components in individual axes.

The underlying graph uses six-neighbor face adjacency. Enumerating
$c\to c+\delta$ with $\delta\in\Delta^+=\{e_x,e_y,e_z\}$ selects one
orientation for each leaf contact. Distinct blocks $A$ and $B$ are
eligible when such a contact joins $G_A$ to $G_B$.
Their candidate is $(a,b,r)$, where
$r=\alpha_B-\alpha_A$. The unit vector $\delta$ identifies a leaf
contact direction, $r$ locates one anchor relative to the other, and
$d$ locates a leaf within its pattern. In particular, the anchors of
face-adjacent blocks need not themselves be face adjacent.
Figure~\ref{fig:spac_dataflow} illustrates how these spatial objects
are merged and expanded.

\subsection{Learning, Encoding, and Decoding}
\label{app:3dbpe_algorithm}

The following paragraphs expand the learning, encoding, and decoding
operations of Algorithm~\ref{alg:3d-bpe-train}.

\para{Initialization and Candidate Frequencies.}
\textsc{InitBlocks} creates a singleton block $(k,c,\{c\})$ for
each input leaf $(c,k)$. Let $\mathcal C_i$ be the multiset of
directed face-adjacent occurrences enumerated for shape $i$ by
\textsc{CountFacePairs}. An occurrence has endpoint blocks $A,B$
and candidate $p(A,B)=(a,b,\alpha_B-\alpha_A)$. The corpus frequency is
\begin{equation}
f(p)=\sum_i\sum_{(A,B)\in\mathcal C_i}
\mathbf 1\{p(A,B)=p\}.
\label{eq:app_candidate_frequency}
\end{equation}
Repeated occurrences contribute to this count. Frequencies are
recomputed on the current blocks at each learning round. The
highest-frequency candidate is selected, with lexicographic
tie-breaking. Learning stops when no candidates remain, the selected
frequency is below $\tau$, or $m$ rules have been learned.

\para{Pattern Construction and Anchor Normalization.}
Lines~\ref{alg:bpe-pattern-start}--\ref{alg:bpe-pattern-end}
express both child patterns relative to $\alpha_A$, then translate
them to the new anchor:
\begin{equation}
Q=\mathcal P_a\cup\{(r+d,k):(d,k)\in\mathcal P_b\},\quad
s=\min_{\mathrm{lex}}\{d:(d,k)\in Q\},\quad
\mathcal P_u=\{(d-s,k):(d,k)\in Q\}.
\label{eq:app_pattern_normalization}
\end{equation}
The new block has anchor $\alpha_A+s$ and support $G_A\cup G_B$.
The pattern is shared by every occurrence of $(a,b,r)$; its
construction depends on relative displacement and leaf IDs.

\para{Matching and Non-Overlapping Replacement.}
Lines~\ref{alg:bpe-replace-start}--\ref{alg:bpe-replace-end}
apply the selected rule to each shape. \textsc{MatchPairs} lists
matching directed block pairs using the shared traversal convention.
The used-block set $\mathcal U$ starts empty for that rule and shape.
A pair is accepted only when neither endpoint has been used;
after replacement, both endpoints enter $\mathcal U$.
This permits each current block to participate in at most one
replacement during the traversal. Each accepted pair is replaced
by $(u,\alpha_A+s,G_A\cup G_B)$, and the ordered rule
$(u,a,b,r)$ is stored for subsequent encoding.

\para{Encoding with the Fixed Table.}
Starting at line~\ref{alg:bpe-encode-start}, a new shape is
initialized as singleton blocks and processed in learned rule order.
\textsc{ApplyRule} uses the stored output ID and pattern and performs
the same matching and non-overlapping replacements.
The resulting sequence is
\begin{equation}
\mathcal S=\operatorname{Sort}_{m(\alpha)}
\bigl[(m(\alpha),u):(u,\alpha,G)\in\mathcal B\bigr].
\label{eq:app_macro_sequence}
\end{equation}
Here $m(\cdot)$ is the Morton map in Eq.~\ref{eq:morton}, whereas
$m$ without an argument is the merge budget. Morton ordering is
applied after spatial merging. The five-bit coordinates map each
anchor to one of $32^3=32768$ Morton indices.

\para{Decoding.}
For every $(z,u)\in\mathcal S$, inverse Morton decoding recovers
$\alpha=m^{-1}(z)$, and $\mathcal P_u$ supplies all pairs
$(\alpha+d,k)$. Taking their union recovers the coordinate--leaf
mapping. Because each stored pattern is expressed directly in
leaf IDs and offsets, the expansion applies to both macro tokens
and singleton leaf tokens.

\subsection{Exact Recovery and Connectivity}
\label{app:3dbpe_reversibility}

\newtheorem{spacproposition}{Proposition}[section]
\begin{spacproposition}[Exact recovery of the sparse mapping]
\label{prop:app_bpe_recovery}
Let $\mathcal M$ have unique coordinates in the legal latent domain.
Suppose the current blocks form a disjoint partition of its leaves,
merges combine disjoint blocks using Eq.~\ref{eq:app_pattern_normalization},
encoding and decoding use the same pattern table, and $m^{-1}$
is the inverse of $m$ on legal anchor coordinates. Then
\begin{equation}
\operatorname{Decode}(\operatorname{Encode}(\mathcal M))
=\mathcal M.
\label{eq:app_bpe_exact_recovery}
\end{equation}
\end{spacproposition}

\begin{proof}
The invariant is that each block expands to exactly its constituent
coordinate--leaf-ID pairs, and the disjoint union of all block
expansions equals $\mathcal M$. Singleton initialization satisfies
this invariant. Consider a merge of $A$ and $B$ with
$r=\alpha_B-\alpha_A$. For a leaf $(d,k)\in\mathcal P_a$,
the normalized offset is $d-s$, and
\begin{equation}
(\alpha_A+s)+(d-s)=\alpha_A+d.
\end{equation}
For a leaf $(d,k)\in\mathcal P_b$, its new offset is $r+d-s$, so
\begin{equation}
(\alpha_A+s)+(r+d-s)=\alpha_B+d.
\end{equation}
Both identities preserve the leaf coordinate and its ID.
Replacing the two disjoint supports by their union therefore
preserves the invariant. Induction covers every accepted merge.
Sorting changes only the order of blocks; inverse Morton decoding
recovers their anchors. Expanding the final blocks thus returns
exactly $\mathcal M$.
\end{proof}

The recovered object is the quantized sparse mapping. Its leaf IDs
select codebook vectors for SDF reconstruction, with surface detail
determined by SDF sampling, quantization, and the VQ-VAE decoder.

\para{Connectivity.}
Each singleton support is face connected. Every accepted merge
joins two face-connected supports with at least one face contact,
so their union remains face connected. By induction, all macro-token
supports are face connected throughout learning and encoding.

\begin{figure}[!htb]
\centering
\includegraphics[width=\linewidth]{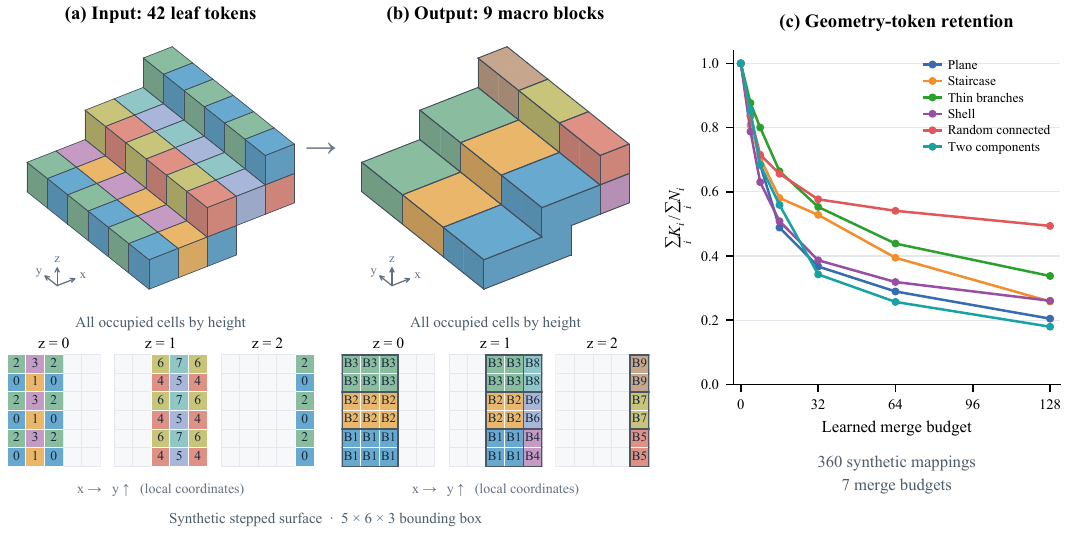}
\caption{\textbf{Synthetic validation of a reference 3D-BPE implementation.}
(a) The 42 leaves of a synthetic three-level stepped surface, colored by
leaf ID. (b) Its nine blocks after applying 128 learned rules; colors and
B1--B9 identify block instances. The matched orthographic views use equal
coordinate scales. The layer maps below show every occupied cell and its
leaf ID or block label, with coordinates translated to a common local origin.
(c) Geometry-token retention $\sum_i K_i/\sum_i N_i$ for 60 mappings
per support family; translated copies are excluded from these ratios.
All 5,040 encode--decode checks, including translated copies and
seven rule budgets, recover the exact sparse mappings.}
\label{fig:appendix_bpe_validation}
\end{figure}

\subsection{Local Information as a Compression Motivation}
\label{app:3dbpe_information}

For a location $c$, let $X_c$ be its leaf ID and
$X_{\mathcal N_k(c)}$ the neighboring IDs defined in
Eq.~\ref{eq:block_neighborhood}. With base-two logarithms, the
optimal marginal predictor has expected log loss $H(X_c)$ bits,
while the optimal predictor conditioned on this neighborhood has
loss $H(X_c\mid X_{\mathcal N_k(c)})$. Their difference is
\begin{equation}
\Delta_{\mathrm{uni}}
=H(X_c)-H(X_c\mid X_{\mathcal N_k(c)})
=I(X_c;X_{\mathcal N_k(c)}).
\label{eq:app_local_information}
\end{equation}
This identity gives a predictive interpretation of the unigram gap:
local geometry supplies information beyond the marginal frequency
of a leaf ID. Grouping recurring connected patterns makes that
structure part of a macro-token type. The empirical boundary
diagnostics in Sec.~\ref{app:residual} then describe the dependencies
remaining between the resulting spatial blocks in each sparse mapping.

\subsection{Synthetic Validation of the Learned Merge and Expansion Rules}
\label{app:3dbpe_synthetic}

An independent reference implementation tests synthetic mappings with eight leaf IDs on a $32^3$ lattice. Supports include planes, staircases, thin branches, box shells, random connected sets, and disconnected patch pairs. IDs follow constant, periodic planar, or three-axis parity patterns. Training uses 96 mappings, 16 per family, threshold $\tau=2$, and up to 128 rules. Encoding tests 360 additional mappings, 60 per family, at budgets $0,4,8,16,32,64,$ and $128$.

Each distinct directed block pair counts once per shape regardless of contact multiplicity. Frequency ties and matching order $(\alpha_A,\alpha_B,a,b)$ are lexicographic, with used endpoints skipped. Every merge checks exact coordinate--ID expansion, disjoint face-connected supports, and lexicographically minimal anchors.

At each budget, original mappings and translations by $(1,2,3)$ yield 5,040 exact round trips. All 23,706 partition and connectivity checks pass. All 2,520 translation comparisons preserve shifted block partitions, with Morton order changing in 2,067 cases. Targeted tests cover negative offsets, moved anchors, multiple contacts, and overlapping candidates. Morton coding round trips cover all 32,768 latent coordinates in the full $32^3$ lattice.

Figure~\ref{fig:appendix_bpe_validation} shows a three-level stepped surface, its complete layers, and per-family retention. Across 360 original mappings, retention falls from $1.000$ at zero rules to $0.570$ at 16 and $0.289$ at 128, yielding 4,085 blocks from 14,152 leaves. These synthetic checks use the stated conventions. Table~\ref{tab:3d_bpe_baselines} separately reports learned SPAC statistics.

\section{Baselines and Additional Analysis of Compression and Dependencies}
\label{app:compression}

\subsection{Formal 1D Baselines}
\label{app:1dbaselines}

The 1D baselines construct merge candidates from consecutive token IDs after sorting the sparse leaves. Let sample $i$ contain $N_i$ leaves $\{(c_{ij},X_{ij})\}_{j=1}^{N_i}$. For a sort key $\kappa$, its initial sequence is
\begin{align}
\pi_i^\kappa
&=\operatorname{argsort}_{j\in\{1,\ldots,N_i\}}
  \kappa_i(c_{ij}),\notag\\*
S_i^\kappa
&=\bigl(X_{i,\pi_i^\kappa(1)},\ldots,
         X_{i,\pi_i^\kappa(N_i)}\bigr).
\label{eq:1d_baseline_sequence}
\end{align}

\Needspace{9\baselineskip}
\para{Morton and Hilbert Ordering.}
Morton ordering interleaves the coordinate bits:
\begin{equation}
\kappa_{\mathrm M}(c)=m_L(c)=\sum_{\ell=0}^{L-1}\Bigl[
    (\lfloor c_x/2^\ell\rfloor\bmod2)2^{3\ell}
  + (\lfloor c_y/2^\ell\rfloor\bmod2)2^{3\ell+1}
  + (\lfloor c_z/2^\ell\rfloor\bmod2)2^{3\ell+2}
\Bigr].
\label{eq:baseline_morton_key}
\end{equation}
Here $L$ is the bit-width of a latent coordinate; $L=5$ for the $32^3$ lattice. The Hilbert baseline uses the 3D Hilbert traversal index:
\begin{equation}
\kappa_{\mathrm H}(c)=h_L(c),\quad
h_L:\{0,\ldots,2^L-1\}^3\longrightarrow\{0,\ldots,2^{3L}-1\}.
\label{eq:baseline_hilbert_key}
\end{equation}

\para{Random Ordering and Rule Selection.}
Random ordering assigns each sample a permutation,
\begin{equation}
\eta_i\sim\operatorname{Unif}(\mathfrak S_{N_i}),\quad
\kappa_{\mathrm R,i}(c_{ij})=\eta_i(j),
\label{eq:baseline_random_key}
\end{equation}
so sequence neighbors are determined by that permutation. At merge round $\ell$, candidate occurrences are the adjacent positions in the current sequences:
\begin{equation}
\mathcal A_{\kappa,\ell}^{1D}
=\bigl\{\!\bigl\{\bigl(S_{i,\ell}^{\kappa}[t],S_{i,\ell}^{\kappa}[t+1]\bigr):
  \;i,\ 1\leq t<|S_{i,\ell}^{\kappa}|\bigr\}\!\bigr\}.
\label{eq:1d_bpe_candidates}
\end{equation}
The multiset retains repeated occurrences. Frequency-based selection chooses
\begin{equation}
(a_\ell,b_\ell)
=\operatorname*{arg\,max}_{(a,b):f_{\kappa,\ell}(a,b)\geq\tau}
f_{\kappa,\ell}(a,b),
\label{eq:1d_greedy_merge}
\end{equation}
where $f_{\kappa,\ell}(a,b)$ is the occurrence count and $\tau$ is the eligibility threshold. The selected rule replaces non-overlapping matches. Learning stops at the merge budget or when no eligible pair remains. Random ordering can also be combined with uniform rule selection,
\begin{equation}
(a_\ell,b_\ell)\sim\operatorname{Unif}
\bigl\{(a,b):f_{\kappa_{\mathrm R},\ell}(a,b)\geq\tau\bigr\},
\label{eq:1d_random_merge}
\end{equation}
which randomizes the eligible rule in addition to the initial ordering of the input leaf sequence.

A 1D rule stores $(a,b)$ and may join spatially separated leaves. A 3D-BPE rule stores $(a,b,r)$ and joins face-connected supports. Table~\ref{tab:3d_bpe_baselines} compares all four tokenizers at $8{,}192$ rules, while Table~\ref{tab:spac_ablation} compares unmerged leaves, Morton 1D-BPE, and SPAC downstream.

\subsection{Geometry Compression and Full Sequence Lengths of LLM Inputs}
\label{app:sequence_lengths}

Let $N_i$ and $K_i$ denote the leaf count and the number of blocks after merging sample $i$, including any remaining singleton blocks. The retained geometry-token fraction is
\begin{equation}
R_{\mathrm{len}}=\frac{\sum_iK_i}{\sum_iN_i}.
\label{eq:app_length_ratio}
\end{equation}
This aggregate ratio is a leaf-count-weighted mean of $K_i/N_i$. Table~\ref{tab:3d_bpe_baselines} gives $0.754$ for 3D-BPE, equivalent to $75.4\%$ of the original geometry-token count, versus $0.803$, $0.820$, and $0.998$ for Morton, Hilbert, and Random 1D.

A complete LLM reference sequence also contains coordinates, text, and control tokens. Writing these four counts for record $i$ as $L_i^{\mathrm{geom}}$, $L_i^{\mathrm{coord}}$, $L_i^{\mathrm{text}}$, and $L_i^{\mathrm{ctrl}}$ gives
\begin{equation}
L_i=L_i^{\mathrm{geom}}+L_i^{\mathrm{coord}}
    +L_i^{\mathrm{text}}+L_i^{\mathrm{ctrl}}.
\label{eq:app_full_length}
\end{equation}
Lengths precede truncation or packing. Across $43{,}180$ text-to-3D records, Leaf and 3D-BPE means are $6{,}873.2$ and $3{,}753.1$, a $45.40\%$ reduction, with 95th percentiles $14{,}675.1$ and $8{,}483.2$. Morton, Hilbert, and Random means are $5{,}443.2$, $5{,}385.4$, and $5{,}645.6$. Full lengths measure serialized records, while $R_{\mathrm{len}}$ measures aggregate geometry counts across the evaluated mappings.

SPAC stores each block as a Morton anchor and geometry token, with leaf patterns in the fixed vocabulary, as in Sec.~\ref{sec:morton_llm}. Table~\ref{tab:3d_bpe_baselines} includes coordinate, formatting, and text costs in full lengths, separately from geometry statistics.

\subsection{Entropy and Boundary Dependency Diagnostics}
\label{app:residual}

\para{Directed Leaf Edges.}
The multiset $\mathcal E_X$ in Eq.~\eqref{eq:leaf_edges} visits each occupied contact $(c,c+\delta)$ once for $\delta\in\Delta^+=\{e_x,e_y,e_z\}$, retaining repeated ID triples. Counts $n_X(a,b,\delta)$ define
\begin{equation}
\hat p_X(a,b,\delta)
=\frac{n_X(a,b,\delta)}{|\mathcal E_X|}.
\label{eq:app_edge_distribution}
\end{equation}
Marginals are obtained by summing these frequencies. The adjacency entropy is the joint entropy
\begin{equation}
\widehat H_2^{3D}(X)=\widehat H(X_u,X_v,\delta),
\label{eq:app_joint_edge_entropy}
\end{equation}
and the directional dependency is
\begin{equation}
\widehat I_X
=\widehat I((X_u,\delta);X_v)
=\widehat H(X_u,\delta)+\widehat H(X_v)-\widehat H(X_u,X_v,\delta).
\label{eq:app_leaf_mi}
\end{equation}
Entropies use base-$2$ logarithms and the convention $0\log_2 0=0$; their units are bits. In this joint MI, the source token and the associated spatial contact direction together form the first random variable.

\para{Macro-Token Boundaries.}
Contacts between distinct blocks form $\mathcal E_B^\partial$, with endpoints relabeled by macro IDs. Each leaf contact counts separately, weighting blocks sharing multiple faces accordingly. Normalizing by $|\mathcal E_B^\partial|$ gives $\hat p_B^\partial(a,b,\delta)$ and MI
\begin{equation}
\widehat I_B^\partial
=\widehat I((B_u,\delta);B_v)
=\widehat H(B_u,\delta)+\widehat H(B_v)-\widehat H(B_u,B_v,\delta).
\label{eq:app_boundary_mi}
\end{equation}
For nonempty edge multisets and $\widehat I_X>0$, the ratios in Eq.~\eqref{eq:bpe_metrics} compare the per-edge MI and the remaining number of edges:
\begin{equation}
R_{\mathrm{MI}}=\frac{\widehat I_B^\partial}{\widehat I_X},
\qquad
R_{\mathrm{edge}}=\frac{|\mathcal E_B^\partial|}{|\mathcal E_X|}.
\end{equation}
Their product is the edge-weighted dependency ratio,
\begin{equation}
R_{\mathrm{wMI}}
=\frac{|\mathcal E_B^\partial|\widehat I_B^\partial}
       {|\mathcal E_X|\widehat I_X}
=R_{\mathrm{MI}}R_{\mathrm{edge}}.
\label{eq:residual_dependency}
\end{equation}
The ratios measure dependency per boundary edge, retained contacts, and their combined change.

\para{Token and Sequence Statistics.}
The macro-token unigram entropy $\widehat H_1(B)$ measures the empirical ID distribution. Multiplying by $R_{\mathrm{len}}$ expresses its cost per original leaf:
\begin{equation}
\widehat H_1(B)R_{\mathrm{len}}
=\frac{\bigl(\sum_iK_i\bigr)\widehat H_1(B)}{\sum_iN_i}.
\label{eq:app_unigram_cost}
\end{equation}
This quantity has units of bits per leaf. The sequence statistics condition on the preceding geometry tokens,
\begin{align}
H_{\mathrm{bi}}(B)&=\widehat H(B_t\mid B_{t-1}),\notag\\*
H_{\mathrm{tri}}(B)&=\widehat H(B_t\mid B_{t-2},B_{t-1}),
\label{eq:app_conditional_entropies}
\end{align}
in bits per predicted token. Unigram cost combines frequency and count, sequence entropies describe short prefixes, and boundary statistics track spatial contacts.

\subsection{Additional Interpretation of the Results}
\label{app:compression_results}

\para{Spatial Adjacency and Merge Budget.}
At $8{,}192$ rules, 3D-BPE leads Table~\ref{tab:3d_bpe_baselines} in retention and unigram cost, $0.754$ and $9.420$ bits per leaf versus Morton at $0.803$ and $9.881$. Its bigram and trigram entropies are lowest at $0.972$ and $0.061$ bits. Hilbert improves trigram entropy over Morton, $0.072$ versus $0.127$, but retains more tokens, $0.820$ versus $0.803$.

Figure~\ref{fig:3d_bpe_sweep} shows retention, normalized unigram cost, and boundary edge fraction decreasing with budget. At high budgets, per-edge MI exceeds the leaf reference, while edge-weighted MI remains below it. The final weighted increase reflects per-edge dependence partly offsetting fewer boundary contacts, as compression removes internal contacts and concentrates surviving dependence.

\para{Downstream Effect of 3D-BPE.}
Table~\ref{tab:spac_ablation} compares leaves, Morton 1D-BPE, and SPAC. From Morton to SPAC, CLIP rises from $23.4$ to $24.4$, FD falls from $26.9$ to $23.7$, and displayed KD falls from $0.27$ to $0.22$. All caption metrics improve, including METEOR from $15.59$ to $21.50$ and SimCSE from $58.04$ to $62.20$, alongside the lowest reported cost in Sec.~\ref{app:efficiency_results}.

\para{Captioning and Generation Comparisons.}
SPAC leads Table~\ref{tab:3d_captioning} with BLEU-1 $22.31$, ROUGE-L $24.43$, METEOR $21.50$, Sentence-BERT $68.24$, and SimCSE $62.20$. The strongest Sentence-BERT and SimCSE baselines score $50.15$ and $50.86$. In Table~\ref{tab:text_image_to_3d}, SPAC achieves FD of $23.7$ and KD of $0.22$, compared with ShapeLLM-Omni at $25.9$ and $0.25$, with KD scaled by $10^2$. Its CLIP score of $24.4$ measures text--shape alignment alongside the distributional metrics, giving a complementary view of generation performance across the evaluated models.

\para{Language Performance after Joint Fine-Tuning.}
In Table~\ref{tab:language_capability}, SPAC raises BBH from $72.5$ to $73.4$. MMLU, MMLU-Pro, and GSM8K score $69.5$, $47.9$, and $86.1$, versus Qwen3-4B at $72.9$, $50.5$, and $87.7$. SPAC leads all compared 3D-capable models on these four benchmarks, with task-dependent differences from the backbone.

\section{Computational Efficiency}
\label{app:efficiency}

\subsection{Sparse Encoder Complexity}
\label{app:encoder_complexity}

For a fixed regular normalized surface of area $A$, grid spacing $h=1/R$ and fixed voxel bandwidth $\eta$ give physical bandwidth $\tau=\eta h$. For sufficiently small $\tau$, dividing band volume $O(A\tau)$ by voxel volume gives
\begin{align}
N_{\mathrm{active}}(R)
&=O\!\left(\frac{A\tau}{h^3}\right)\notag\\*
&=O\!\left(\frac{A\eta h}{h^3}\right)
=O(A\eta R^2).
\label{eq:app_surface_scaling}
\end{align}
Thus, refining the grid at fixed voxel-width bandwidth increases the active input on the scale of a surface.

Let layer $\ell$ have $N_\ell$ active sites, kernel width $k_\ell$, and channels $C_{\ell,\mathrm{in}}$ and $C_{\ell,\mathrm{out}}$. With fixed depth, channels, kernels, batch size, and local neighborhoods, and $N_\ell=O(N_{\mathrm{active}})$ at every scale,
\begin{align}
C_{\mathrm{enc}}(R)
&=\sum_\ell O\!\left(N_\ell k_\ell^3
C_{\ell,\mathrm{in}}C_{\ell,\mathrm{out}}\right)\notag\\*
&\quad+\sum_\ell O(N_\ell)
=O(R^2).
\label{eq:app_encoder_cost}
\end{align}
The second term covers attention with a fixed local neighborhood. Holding the physical bandwidth $\tau$ fixed instead yields $N_{\mathrm{active}}=O(R^3)$.

Here $R$ is the input SDF resolution, with each latent axis downsampled by eight. The subsequent LLM sequence has length $L$ and $O(L^2)$ global attention interactions. Sparse encoding and sequence compression therefore affect different pipeline stages.

\subsection{Measurement Protocol}
\label{app:efficiency_protocol}

Encoders in Fig.~\ref{fig:cmp} run independently on one NVIDIA A100 in full precision with batch size $1$, using the same normalized geometry at each resolution. SPAC fixes voxel bandwidth and local neighborhoods, while baselines use native representations. Surface sampling grows with $R^2$ until capped. Capped points remain plotted but are excluded from slope fitting, with $R=256$ the largest uncapped resolution. Active-input and FLOP slopes are $1.991$ and $2.164$, consistent with surface scaling and finite-resolution and downsampling effects in the sparse encoder.

\para{Four Measurement Scopes.}
Forward FLOPs count convolution, linear, and attention operations with two operations per multiply--accumulate. Forward latency uses synchronized passes after warm-up. Prepared-input time adds device transfer and tensor construction or ordering, excluding mesh loading and preprocessing. Peak VRAM is the maximum allocated inference memory. Figure~\ref{fig:cmp} shows these four scopes on logarithmic axes, with analytical dense FLOPs beyond executable resolutions.

\subsection{Encoder and Downstream Costs}
\label{app:efficiency_results}

\para{Encoder Comparison.}
ShapeLLM-Omni uses a dense 3D VQ-VAE~\citep{ye2025shapellmomni}, PointLLM uses Point-BERT with $512$ groups~\citep{xu2023pointllm}, and EVA01 uses its generation O-Voxel sparse VAE~\citep{yang2026eva01unifiednative3d}. The EVA01 understanding encoder overlaps with PointLLM. These choices explain the trends in Fig.~\ref{fig:cmp}, faster growth for dense encoding, nearly constant PointLLM FLOPs, and higher constant cost for the larger EVA01 sparse encoder in its native configuration.

At $R=256$, SPAC uses $137.98$ G FLOPs and $85.04$ ms, versus $41.38$ T FLOPs and $4{,}227.84$ ms for ShapeLLM-Omni. At low resolutions, ShapeLLM-Omni has lower latency. PointLLM combines lower latency with growing sampling and neighborhood memory. Prepared-input time and VRAM capture tensor construction and intermediates.

\para{Downstream Ablation Costs.}
Table~\ref{tab:spac_ablation} gives $69{,}771.35$ G FLOPs and $348.40$ s without BPE, $56{,}299.82$ G and $262.93$ s for Morton 1D-BPE, and $41{,}122.05$ G and $171.69$ s for SPAC. Relative to Morton, SPAC reduces FLOPs by approximately $27.0\%$ and latency by $34.7\%$, alongside the quality gains in Sec.~\ref{app:compression_results}. Full sequence lengths in Sec.~\ref{app:sequence_lengths} separately measure the LLM sequence burden.

\subsection{Controlled SDF Band Scaling Experiment}
\label{app:band_scaling}

We test Eq.~\eqref{eq:app_surface_scaling} by counting analytic sphere SDF samples on cell-centered grids in $[0,1]^3$, with center $(0.5,0.5,0.5)$ and radius $r_s=0.3$. At each center $\mathbf x$,
\begin{equation}
s(\mathbf x)=\|\mathbf x-(0.5,0.5,0.5)\|_2-r_s.
\end{equation}
We count cells satisfying $|s(\mathbf x)|\leq\tau$ for fixed voxel half-width $\tau=0.5/R$ or fixed physical half-width $\tau=0.5/128$, at resolutions $32$, $48$, $64$, $96$, $128$, $192$, $256$, $384$, and $512$. All cells are evaluated in CPU double precision without sampling caps.

\begin{figure}[!htb]
\centering
\includegraphics[width=0.60\linewidth]{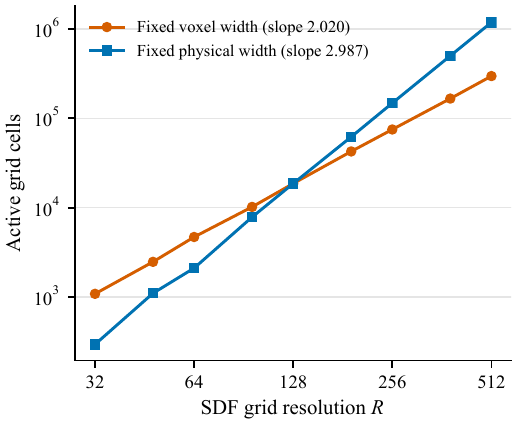}
\caption{\textbf{Active SDF samples for a fixed sphere.} Markers show grid counts for fixed voxel half-width $\tau=0.5/R$ and fixed physical half-width $\tau=0.5/128$; solid segments connect successive resolutions. Log--log slopes fitted to all nine resolutions are $2.020$ and $2.987$, respectively. The curves meet at $R=128$, where the two bandwidths coincide.}
\label{fig:appendix_band_scaling}
\end{figure}

Figure~\ref{fig:appendix_band_scaling} shows the counts. Unweighted least-squares fits of $\log N_{\mathrm{active}}$ against $\log R$ across all nine resolutions give slopes $2.020$ and $2.987$ for fixed voxel and physical widths. Both bands have $18{,}648$ active cells at $R=128$, increasing to $296{,}744$ and $1{,}184{,}368$ at $R=512$. The controlled counts match the predicted quadratic and cubic growth as grid resolution increases.

\section{Limitations and Future Directions}
\label{sec:limit}

\para{Modalities and Outputs.}
SPAC models geometry and text and generates untextured meshes. Image conditioning and appearance require added representations. CLIP~\citep{radford2021learning} could supply image features to condition the LLM alongside the existing sparse geometry-token representation.

\para{Representation and Reconstruction.}
SDF resolution, codebook, and decoder capacity determine geometric detail. Exact 3D-BPE expansion preserves quantized leaves, while reconstruction depends on the tokenizer. Examples G20 and G24 guide future work on surface fidelity, codebook use, and preservation of small structures.

\para{Sequence Length and Computation.}
Sequence length follows the number of active sites and autoregressively predicted coordinate--token pairs. Near-quadratic encoder scaling assumes a fixed regular surface, voxel bandwidth, architecture, and local neighborhoods. Surface complexity, channels, and neighborhood size alter cost.

\para{Data and Generalization.}
The $50$k meshes provide training coverage across their shapes and descriptions. Qualitative examples from these training objects show variation in part coverage and caption attributes. Broader mixed 2D--3D data could improve robustness. Dynamic or 4D coordinate--token maps additionally require temporal representations and consistency.

\Needspace{5\baselineskip}
\section{Ethics and Societal Impact}
\label{sec:impacts}

\para{Broader Impacts.}
Efficient 3D modeling supports gaming, film, virtual and augmented reality, design, education, and cultural preservation. Misuse includes counterfeiting, unauthorized copying, and deceptive or harmful content. Release licenses will prohibit malicious use. We encourage respect for data and design rights, transparent documentation, provenance tracking, watermarking, and safeguards tailored to each application.

\para{Use of Large Language Models.}
LLMs assisted with language editing, literature organization, consistency checks, and routine technical tasks. Authors remain responsible for methodology, experiments, and verification of mathematical claims and results.

\section{Additional Qualitative Results}
\label{app:qualitative}

\subsection{Understanding and Generation Comparisons}
\label{app:qualitative_comparisons}

Figure~\ref{fig:understanding_case_study} compares barrel, house, chest, and tree captions. SPAC identifies house windows, door, and roof, the rectangular chest body and hinged lid, and the tree trunk and canopy; nearby animals are outside the tree caption's coverage. Several other outputs describe only geometric forms, illustrating differences in recognition and scene coverage.

\begin{figure}[p]
\centering
\includegraphics[width=\linewidth]{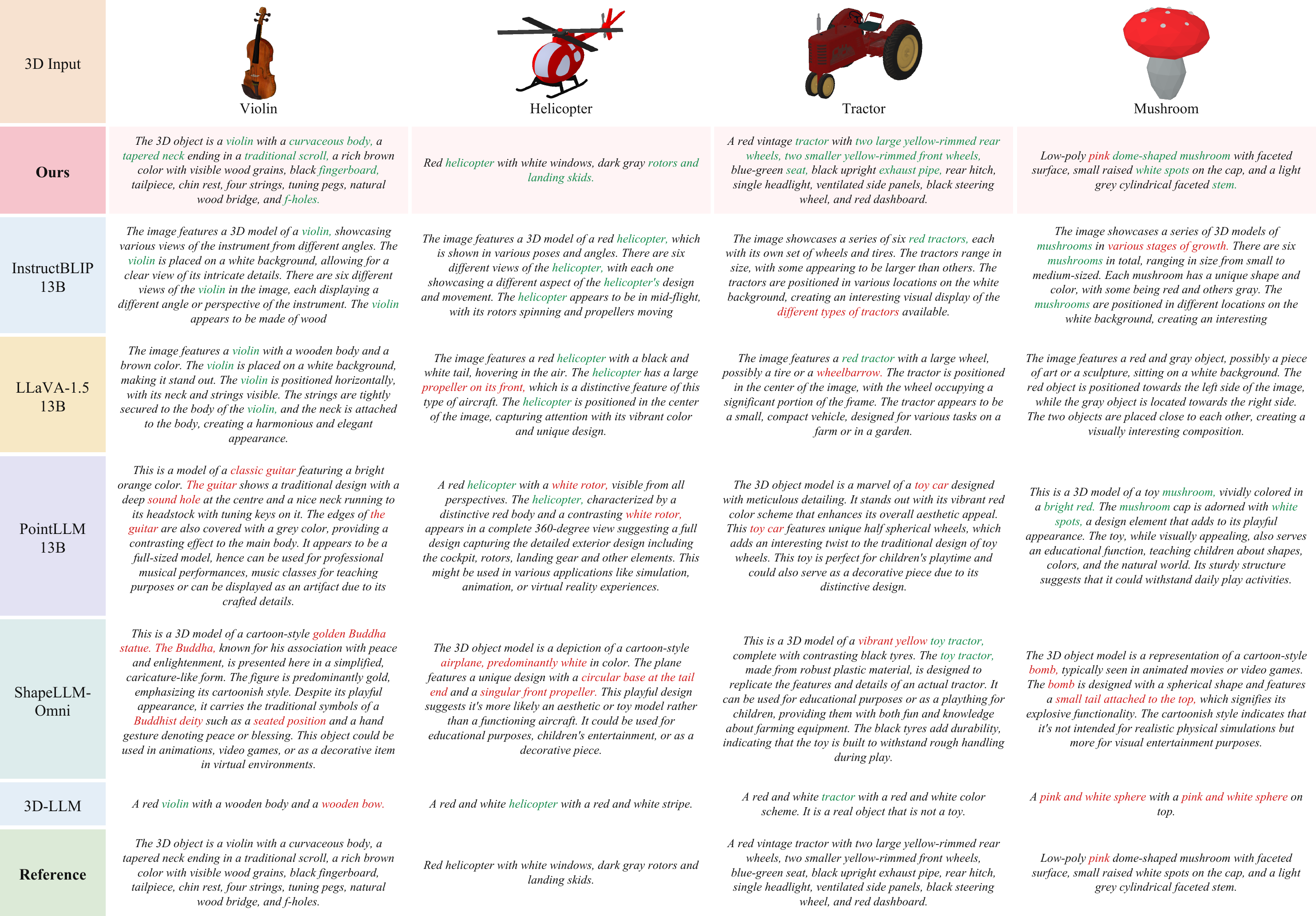}
\caption{\textbf{Additional 3D understanding comparisons.} These are training-seen Toys4K objects. Captions are saved outputs, and Reference denotes dataset descriptions. Green/red phrases highlight supported/mismatched visible attributes.}
\label{fig:understanding_case_study_extra}
\end{figure}

Figure~\ref{fig:understanding_case_study_extra} adds training examples of a violin, helicopter, tractor, and mushroom. Inputs are saved SDF leaf tokens for SPAC, six RGB views for InstructBLIP and LLaVA, surface points for PointLLM, geometry tokens for ShapeLLM-Omni, and stored 3D features for 3D-LLM. SPAC is prompted for one sentence and baselines for detailed captions. For the mushroom, SPAC matches the dataset description of a pink cap, while the rendered example appears red. All captions retain saved outputs.

In Fig.~\ref{fig:generation_case_study}, SPAC retains the requested bulging barrel body, paired butterfly wings, and curved chest lid. Colored renderings combine SPAC geometry with TRELLIS appearance~\citep{xiang2025structured}, following ShapeLLM-Omni~\citep{ye2025shapellmomni}.

\subsection{Extended Generation Examples}
\label{app:qualitative_generation}

\begin{figure}[p]
\centering
\includegraphics[width=\linewidth]{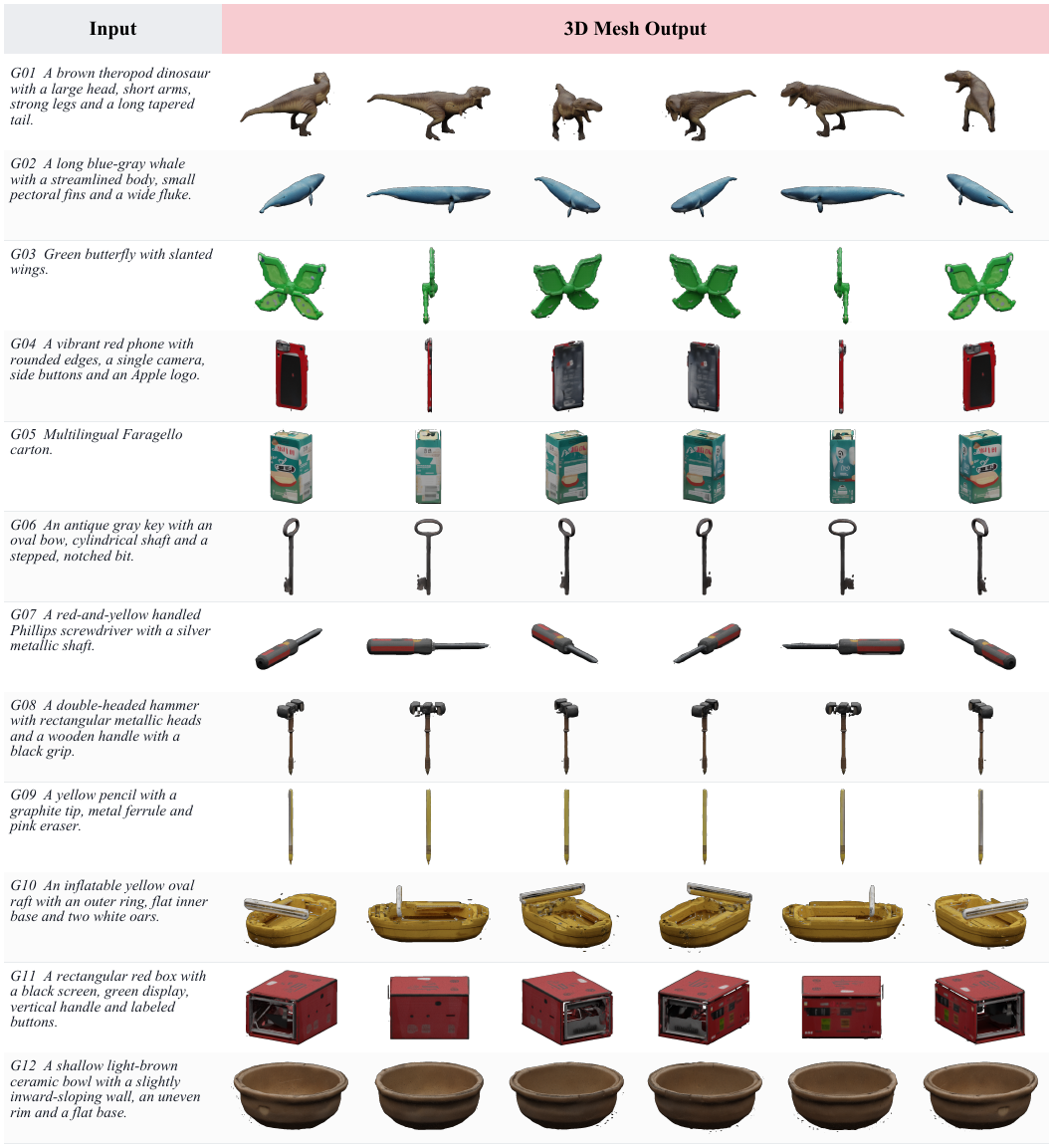}
\caption{\textbf{Additional text-to-3D examples, G01--G12.}}
\label{fig:appendix_generation_01}
\end{figure}

\begin{figure}[p]
\centering
\includegraphics[width=\linewidth]{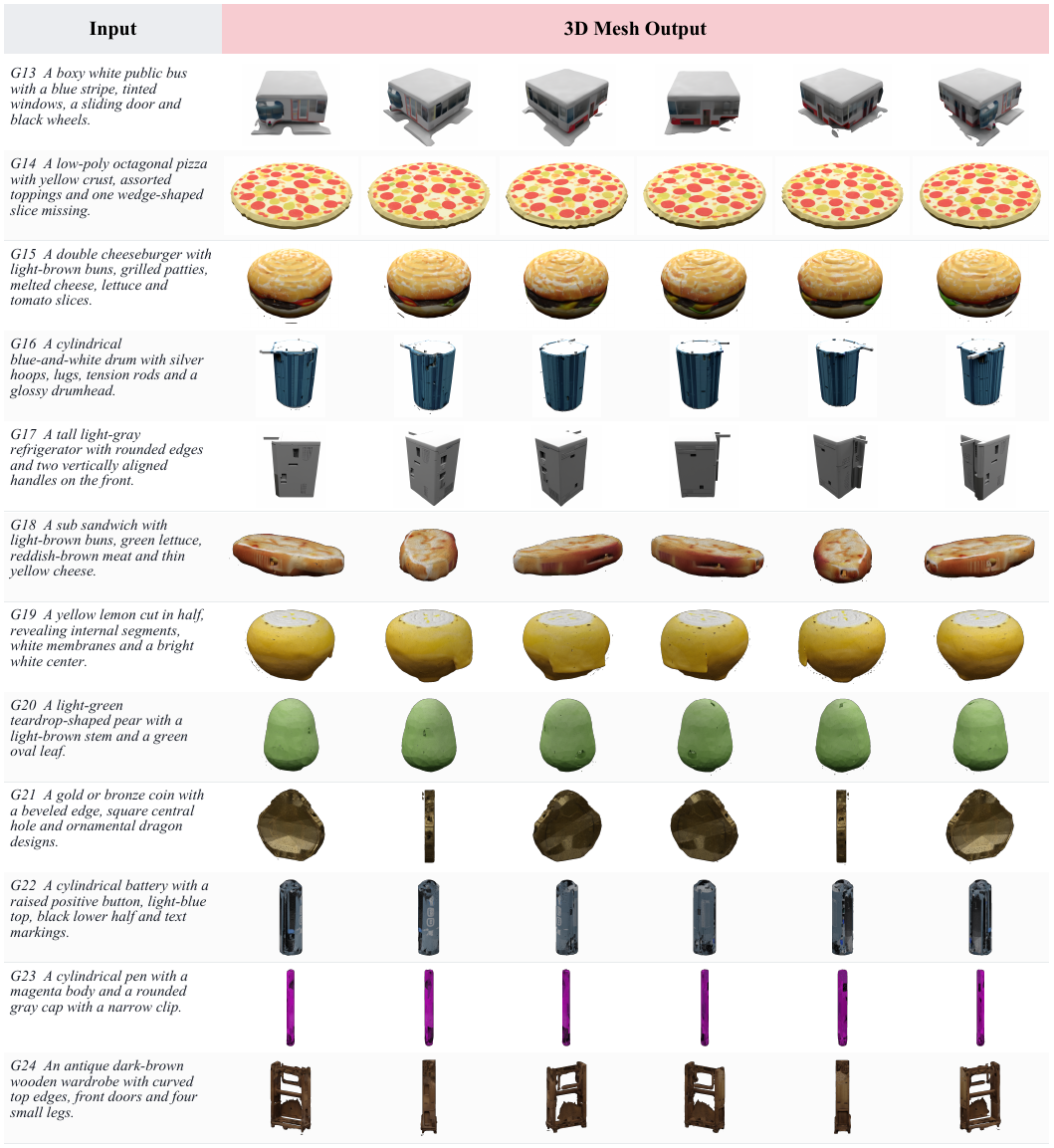}
\caption{\textbf{Additional text-to-3D examples, G13--G24.}}
\label{fig:appendix_generation_02}
\end{figure}

\begin{figure}[p]
\centering
\includegraphics[width=\linewidth]{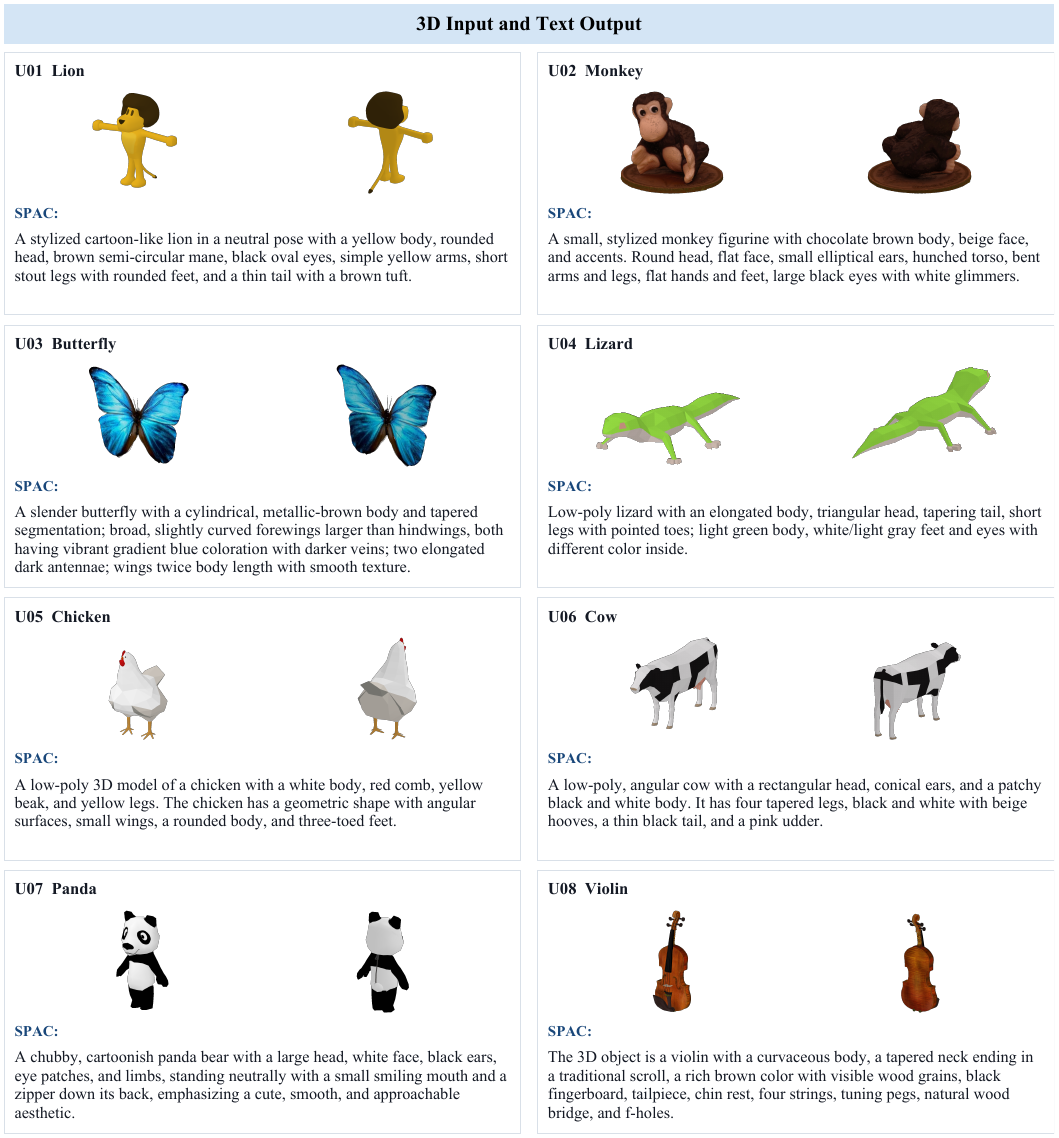}
\caption{\textbf{Additional 3D-to-text examples, U01--U08.}}
\label{fig:appendix_understanding_01}
\end{figure}

\begin{figure}[p]
\centering
\includegraphics[width=\linewidth]{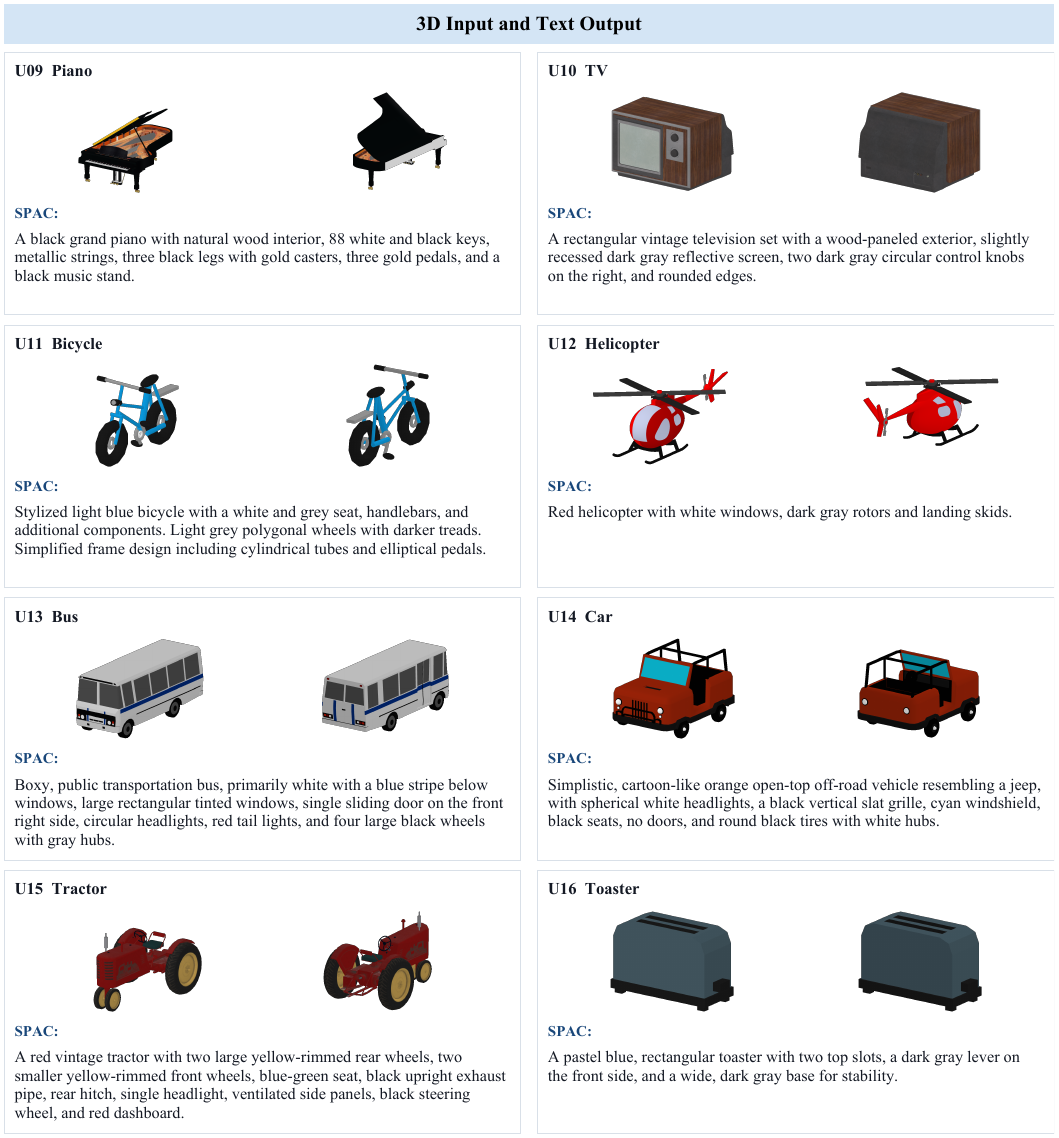}
\caption{\textbf{Additional 3D-to-text examples, U09--U16.}}
\label{fig:appendix_understanding_02}
\end{figure}

\begin{figure}[p]
\centering
\includegraphics[width=\linewidth]{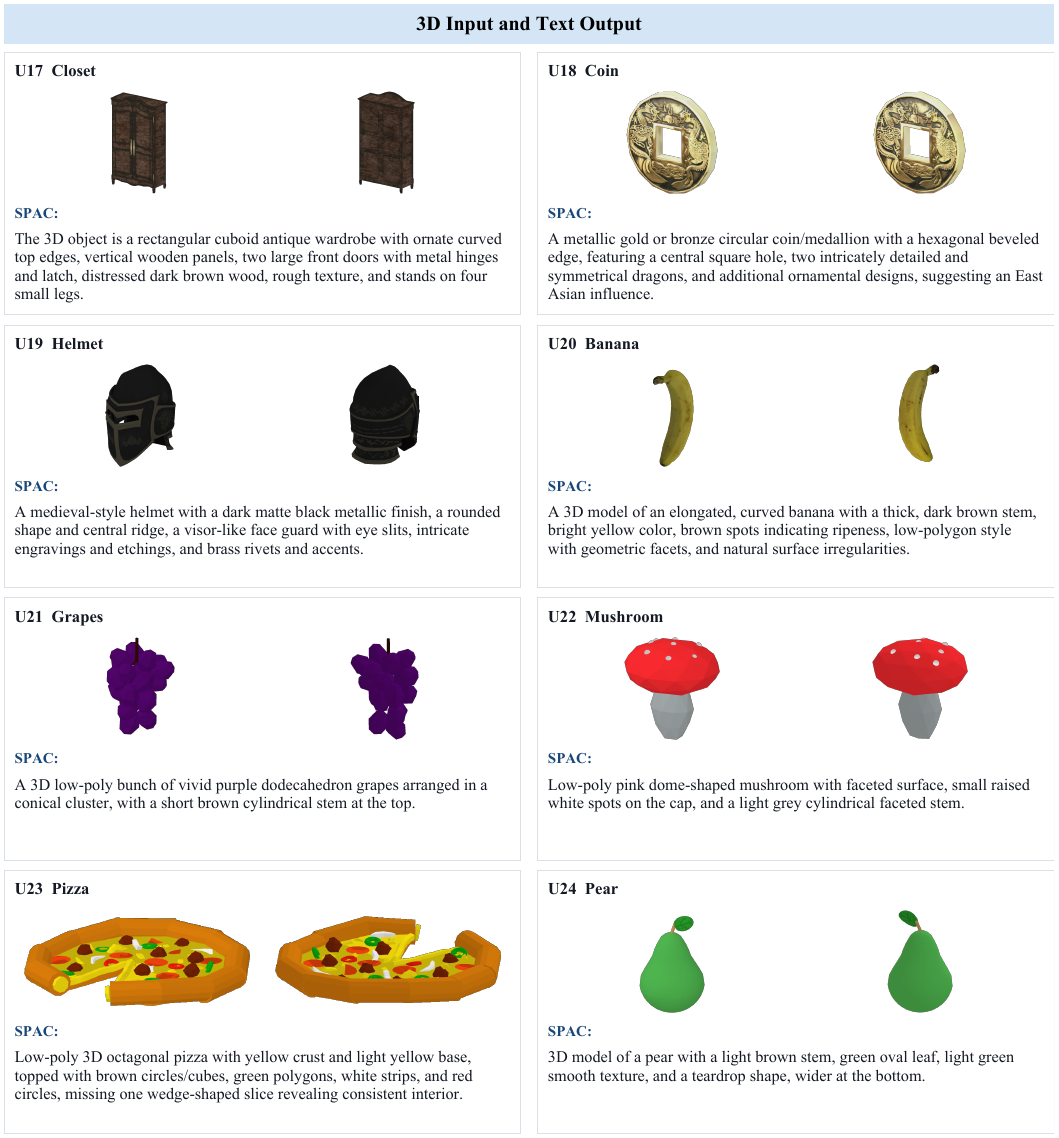}
\caption{\textbf{Additional 3D-to-text examples, U17--U24.}}
\label{fig:appendix_understanding_03}
\end{figure}

Figures~\ref{fig:appendix_generation_01} and~\ref{fig:appendix_generation_02} show G01--G24, and Figs.~\ref{fig:appendix_understanding_01}--\ref{fig:appendix_understanding_03} show U01--U24. Generation prompts are shortened and captions remain verbatim. G13--G17 use saved outputs from an earlier Toys4K SPAC evaluation, with TRELLIS reconstructing geometry and appearance from generated meshes. Other generation examples retain SPAC geometry with TRELLIS appearance. These and U01--U24 use training objects. Understanding figures show original materials, but SPAC receives SDF geometry.

Multiple views show the thin shafts and wider ends of key G06 and hammer G08, and the open interior and raised rim of raft G10. G13--G24 span vehicles, food, a musical instrument, and household objects. Bus G13 has a boxlike body and repeated windows, burger G15 stacked layers, and drum G16 a cylindrical body and circular head. G14 shows a complete disc for a wedge-cut prompt, G20 a pear body without the specified stem and leaf, and G24 an open frame with crosspieces for a wardrobe with front doors. These examples illustrate part-level correspondence between prompts and generated geometry.

\Needspace{10\baselineskip}
\subsection{Extended Understanding Examples}
\label{app:qualitative_understanding}

Captions range from brief categories to part inventories. They describe the elongated body, tapering tail, and short legs of lizard U04, the curved body and narrow neck of violin U08, the rotors and landing skids of helicopter U12, and the large rear wheels, smaller front wheels, seat, and exhaust pipe of tractor U15. U18 identifies the central coin opening and U23 the missing pizza wedge. U09 specifies 88 piano keys, while the displayed views leave the count unresolved. U22 uses the pink-cap description found in the dataset reference. Alongside the tree caption in Fig.~\ref{fig:understanding_case_study}, these examples characterize descriptive specificity, scene coverage, and alignment with rendered appearance.

\par\medskip
\noindent\begin{minipage}{\linewidth}
\centering
\captionsetup{type=figure}
\includegraphics[width=\linewidth]{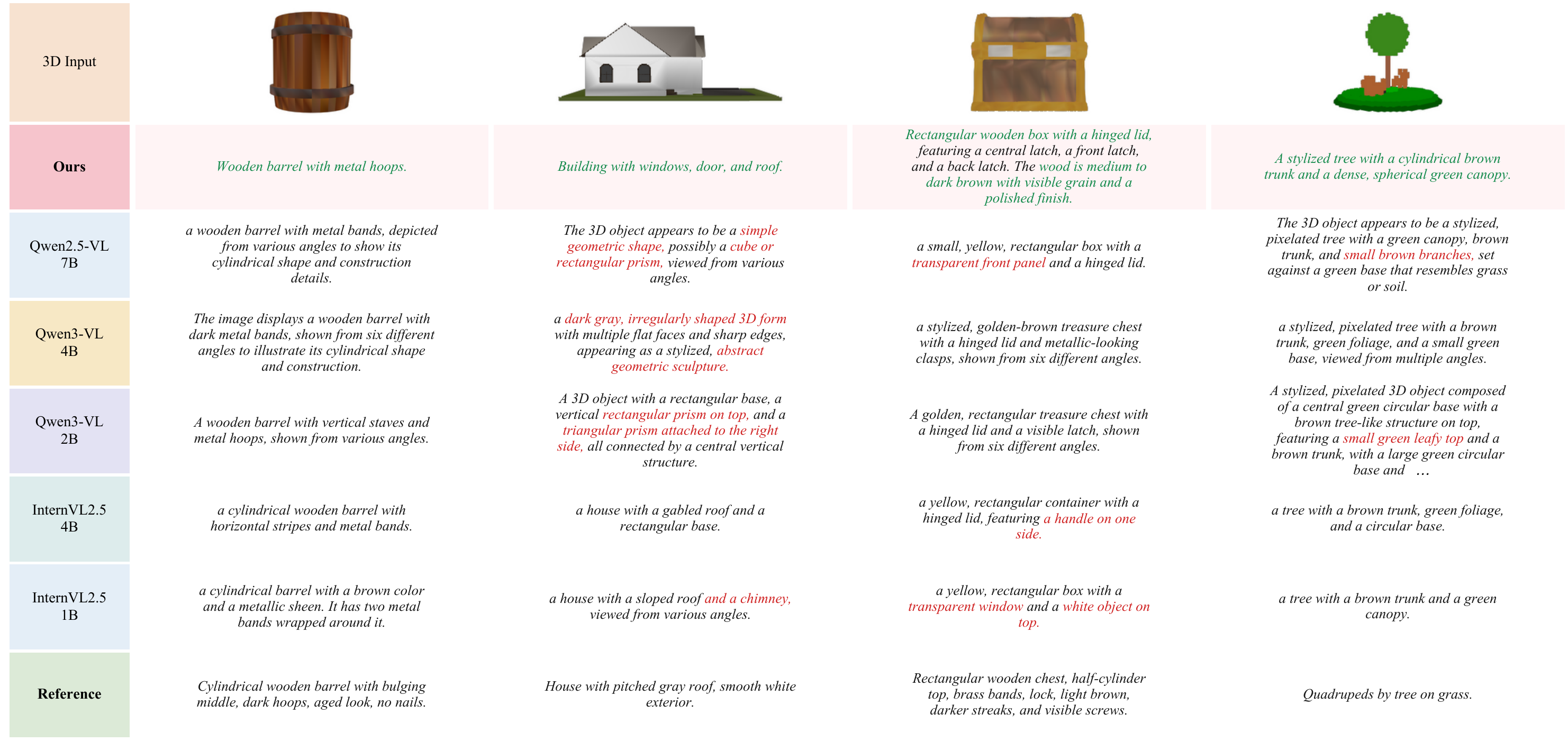}
\caption{\textbf{Qualitative comparisons with other baselines on the 3D understanding task.}}
\label{fig:understanding_case_study}
\end{minipage}\par

\clearpage